\documentclass[journal]{IEEEtran}
\usepackage{cite}

\ifCLASSINFOpdf

\else

\fi

\usepackage{amsmath}
\usepackage{amsfonts}
\usepackage{amssymb}
\usepackage{graphicx}
\usepackage{color}
\usepackage{multirow}
\usepackage{float}

\ifCLASSOPTIONcompsoc
  \usepackage[caption=false,font=normalsize,labelfont=sf,textfont=sf]{subfig}
\else
  \usepackage[caption=false,font=footnotesize]{subfig}
\fi

\usepackage{url}

\begin{document}
%
\title{Learning and Inferring a Driver's Braking Action in Car-Following Scenarios}
\author{Wenshuo~Wang,~\IEEEmembership{Student Member,~IEEE,}
	Junqiang~Xi,
	and Ding~Zhao
\thanks{\textit{Corresponding Author: Junqiang Xi and Ding Zhao}}
	\thanks{W. Wang is with the Department of Mechanical Engineering, Beijing Institute of Technology (BIT), Beijing, China, 100081, and also with the Department of Mechanical Engineering, University of Michigan, Ann Arbor, MI, 48109, USA. He was also with the University of California at Berkeley (UCB), CA, 94720, USA. e-mail: wwsbit@gmail.com}
	\thanks{J. Xi is with the School of Mechanical Engineering, Beijing Institute of Technology (BIT), Beijing, China, 100081. e-mail: xijunqiang@bit.edu.cn}
	\thanks{D. Zhao is an Assistant Research Scientist with the Department of Mechanical Engineering, University of Michigan, Ann Arbor, MI 48109-2150, USA. e-mail: zhaoding@umich.edu}
	}


\maketitle

\begin{abstract}
Accurately predicting and inferring a driver's decision to brake is critical for designing warning systems and avoiding collisions. In this paper we focus on predicting a driver's intent to brake in car-following scenarios from a perception-decision-action perspective according to his/her driving history.  A learning-based inference method, using onboard data from CAN-Bus, radar and cameras as explanatory variables, is introduced to infer drivers' braking decisions by combining a Gaussian mixture model (GMM) with a hidden Markov model (HMM). The GMM is used to model stochastic relationships among variables, while the HMM is applied to infer drivers' braking actions based on the GMM. Real-case driving data from 49 drivers (more than three years' driving data per driver on average) have been collected from the University of Michigan Safety Pilot Model Deployment database. We compare the GMM-HMM method to a support vector machine (SVM) method and an SVM-Bayesian filtering method.  The experimental results are evaluated by employing three performance metrics: \textit{\textbf{accuracy}},  \textit{\textbf{sensitivity}}, and \textit{\textbf{specificity}}. The comparison results show that the GMM-HMM obtains the best performance, with an accuracy of 90\%, sensitivity of 84\%, and specificity of 97\%.  Thus, we believe that this method has great potential for real-world active safety systems. 
%
%
%
\end{abstract}


\IEEEpeerreviewmaketitle

\section{Introduction}

\subsection{Motivation}

\IEEEPARstart{P}{redicting} and inferring drivers' braking actions in advance are critical for avoiding collisions in car-following scenarios. The National Highway Traffic Safety Administration (NHTSA) reported that rear-end collisions accounted for 32.4\% of a total of 1,966,000 crashes in the United States in 2014 \cite{NHTSA}. To prevent the rear-end collisions, a wide variety of forward collision avoidance systems have been developed, such as forward collision warning (FCW) systems \cite{wang2016forward,takada2014effectiveness},  pre-crash brake assist (PBA) systems, and autonomous emergency braking (AEB) systems \cite{guononlinear,AlbertoCertainty,savino2016robust}.

\begin{figure}[t]
	\centering
	\includegraphics[width = 0.48\textwidth]{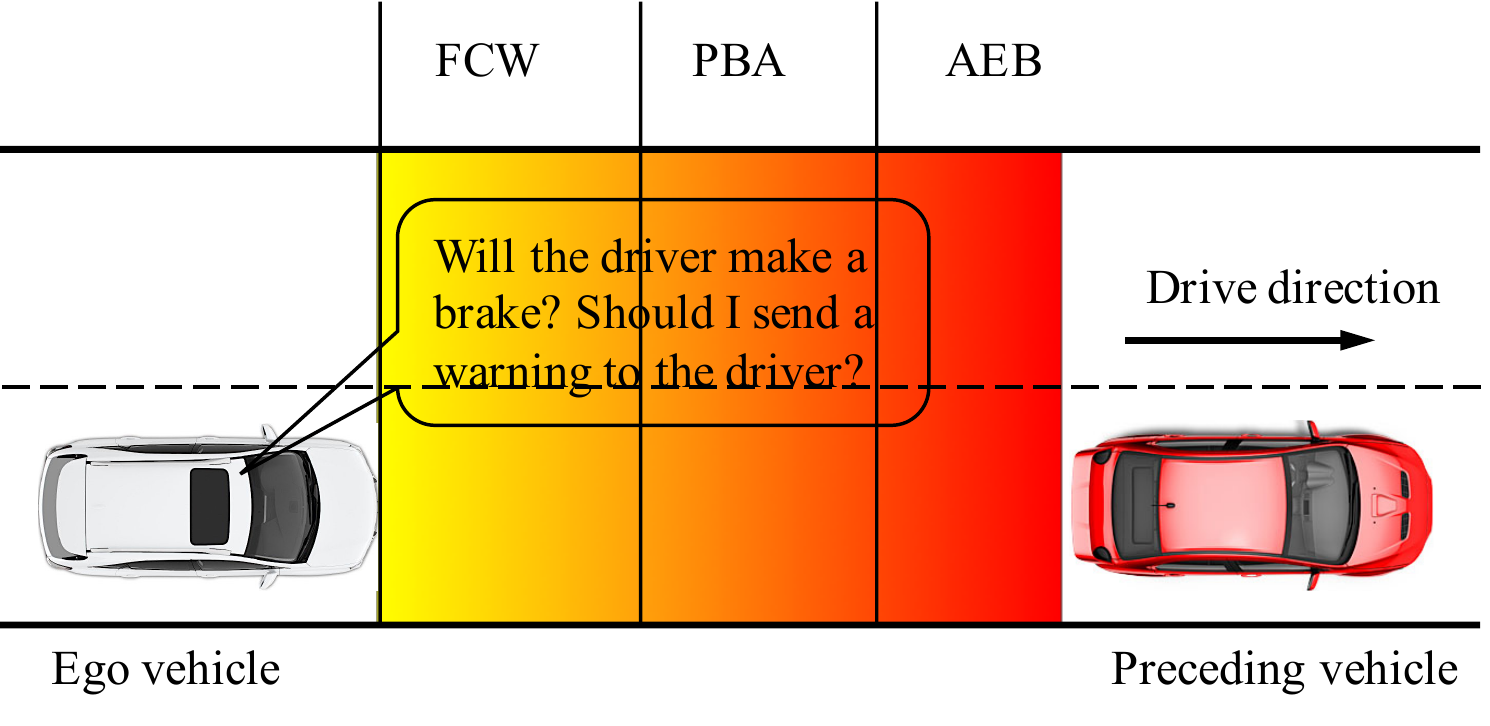}
	\caption{The forward collision avoidance systems for car-following scenarios.  The red (dark) area behind the preceding vehicle represents the critical region where braking actions are required. The yellow (lighter) area represents the region in which the driver should be warned if unaware of the situation.}
	\label{figure1}
\end{figure}

The FCW system is an active safety device that warns drivers by a visual, audio, or tactile means when a potential collision is detected \cite{ahtamad2016warning,meng2015tactile}. FCW systems have been proven to have a positive effect on improving traffic safety. A well-known problem with the FCW systems, however, is false-positive alarms.  Sometimes, the false alarm rate is so high that it reduces its acceptability to end-users.
One key to reducing false-positive alarms is to \textit{infer}\footnote[1]{In this paper, \textit{inferring} means that deducing drivers' braking actions using the model learned from their historical data.} and accurately predict drivers' braking actions and decide correctly whether to deliver a warning to the drivers. As shown in Fig. \ref{figure1}, the FCW system that can correctly judge ``\textit{Will the driver brake?}'' at the next time step and determine ``\textit{Should I send a warning to the driver?}'' can be more attractive to end-users. If it is inferred that the driver will not brake, the FCW systems should then send a warning to the driver so as to avoid a collision \cite{mccall2007driver}. 
\subsection{Related Research}
\begin{figure}[t]
	\centering
	\subfloat[Methods based on foot gesture/movement data.]{\includegraphics[width = 0.48\textwidth]{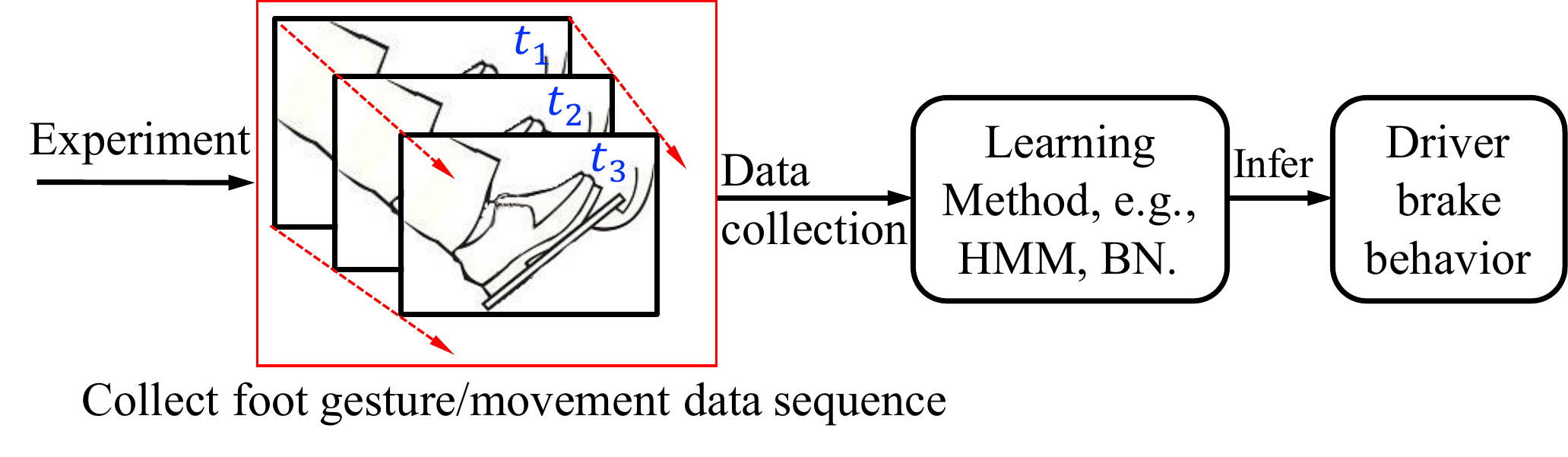}}\\
	\subfloat[Methods based on perception-decision-action (PDA) data.]{\includegraphics[width = 0.48\textwidth]{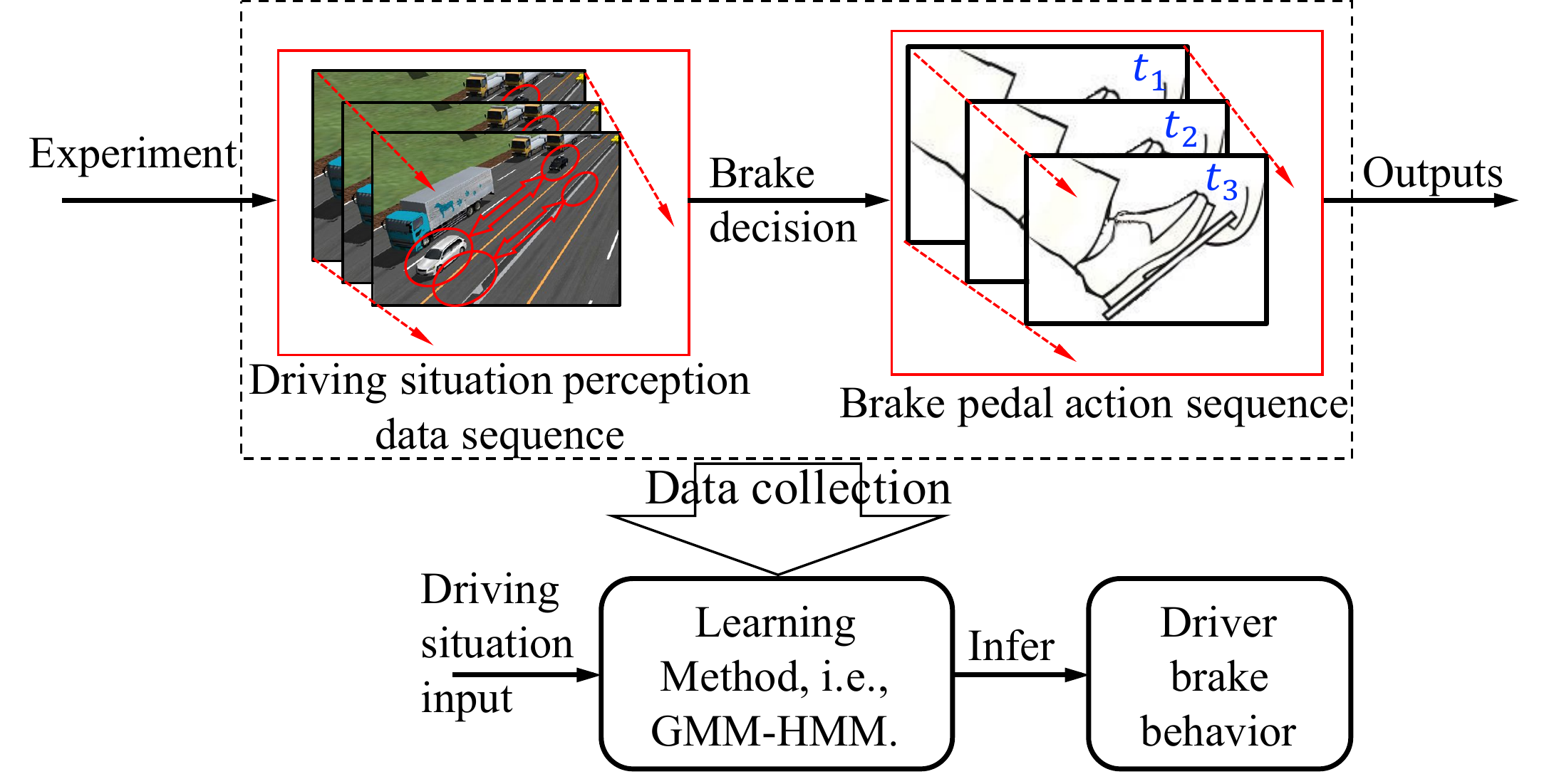}}
	\caption{Comparisons between (a) the foot gesture/movement-based methods  and (b) the main idea of our developed method.}
	\label{Comp_methods}
\end{figure}

Generally, drivers perform braking behavior through two succeeding stages, i.e., decision-making and decision execution. The decision-making stage is reflected by the question ``\textit{Will the driver brake?}'' and the execution stage is reflected by the question ``\textit{What kind of brake style will the driver prefer?}''. In this research, we focus on the first stage, which is the preconditioned to execute the decision and can be found in many existing literatures. For example, Tran \textit{et al}. \cite{tran2012modeling} predicted driver foot behavior under Stop-\&-Go conditions based on the camera data using a hidden Markov model (HMM). The driver foot behavior was decomposed into seven states to characterize the behavior of engaging the acceleration/brake pedal. The predicted driver's foot behavior was used to reduce the possibility of annoying alarm in collision warning systems\cite{sakabe2002development}, but the data used for training HMM should first be labeled, which was labor-intensive, and the vision-based data of the foot gesture depends heavily upon the light on the foot, i.e., requires flashlight illumination\cite{tran2012modeling}. Besides, the methods based on the foot gesture data can not directly reflect how the driver makes decisions to brake when perceiving the current driving situation, as shown in Fig. \ref{Comp_methods}(a).  Pugeault and Bowden \cite{pugeault2015much} predicted driver braking behavior using a statistical learning approach with vision-based data and reached an accuracy of 80\%. McCall and Trivedi \cite{mccall2006human,mccall2007driver} applied a Bayesian network (BN) framework to predict the need for drivers' braking actions using in-vehicle data and surroundings information in seven dimensions, including steering angle, wheel speed, longitudinal/lateral acceleration, yaw rate, brake pedal pressure, and acceleration pedal position. The limitation of using the BN method, however, is that the probability between that of the driver not intending to begin braking and that of the need to brake is not always conditionally independent. Furthermore, the BN method is computationally expensive, particularly for high-dimensional data. 

In addition to using the vehicle-depended data and the camera data, physiological signals could also provide hints of braking behavior. For example, researchers in \cite{khaliliardali2015action,haufe2014electrophysiology} utilized anticipation-related electroencephalogram (EEG) signals to predict the driver braking behavior in a driving simulator (with performance 0.83$ \pm $0.13) and real-world driving. In \cite{abbink2011measuring}, Abbink \textit{et al}. used haptic feedback to inform and infer drivers' braking actions to support car following by measuring neuromuscular control dynamics of legs with electromyogram (EMG) sensors.  When drivers tend to brake, the measures from their foot and/or brains can indirectly reflect their intents, but could lead to a lag between prediction results and the their braking actions because of stimulus delay. The EEG/EMG data could reflect driver's decision-making in braking, but in a real driving case, human drivers would not like to wear the EEG or EMG signal collection equipment when driving, which strongly limits their applications to real vehicles. Differing from research directly using maneuver and physiological signals, Mulder \textit{et al}. \cite{mulder2010active} predicted the driver's action of hitting/releasing the gas pedal by analyzing the parameters of a linear control-theoretic driver-vehicle model based on a Monte Carlo approach. The prediction results were then applied to develop a haptic gas pedal feedback system. However, the linear control-theoretic driver-vehicle model may be inadequate and inappropriate to describe nonlinear, stochastic, dynamic processes\cite{nechyba1998stochastic} like the decision-making process of human drivers.

Based on the aforementioned discussions, it can be known that directly utilizing the camera-related signals or EEG/EMG signals to predict drivers' braking action without considering the fact that human driver behavior is dynamically changing could impede the in-depth applications to the FCW systems. 

\subsection{Contributions}
Inspired by the fact that human reasoning and decision-making involve probabilistic inferential processes (e.g., Markov processes)\cite{donoso2014foundations}, we introduce a GMM-HMM approach, which combines a Gaussian mixture model (GMM) and an HMM, to \textit{learn}\footnote[2]{In this paper, \textit{learning} means that acquiring knowledge of the dynamic and stochastic process of how drivers take braking actions.} and \textit{infer} drivers' braking actions from the perception-decision-action perspective, as shown in Fig. \ref{Comp_methods}(b).  Compared with other methods (e.g., BN and fuzzy logic), the GMM-HMM method has the following advantages: 
\begin{itemize}
	\item It requires fewer parameters to be estimated and the model parameter can be learned in an off-line phase.
	\item It requires less or no effort to label data. All the data are formulated by a joint probability density function.
	\item Many on-the-shelf estimation techniques (e.g., expectation maximization (EM) algorithm) can be directly used in this model, which makes it be easy for applications.
	\item A joint probability density function directly describes the relationships between variables, rather than assuming the conditional independence  between ``brake'' and ``no brake'' as in \cite{mccall2006human,mccall2007driver}.
\end{itemize}
Using the GMM-HMM method that can capture the underlying stochastic and dynamic characteristics of driver behaviors, we aim to infer drivers' braking actions in car-following scenarios. This work presents the following contributions:
\begin{enumerate}
	\item Unlike other research in \cite{tran2012modeling,pugeault2015much,mccall2006human,mccall2007driver,haufe2014electrophysiology}, drivers' braking action is inferred using the states derived from the ego vehicle and the preceding vehicle (Fig. \ref{Comp_methods}(b)), rather than using the EEG/EMG data or the video/camera information of drivers' foot gestures and movements.
	\item Differing from existing research, a framework is proposed from the perception-decision-action \cite{windridge2013characterizing} perspective for modeling, learning, and inferring drivers' braking actions based on the GMM-HMM method.
\end{enumerate}

Instead of investigating a driver's braking style, this paper mainly focuses on inferring the driver's intent to brake (Fig. \ref{figure1}), which is essential to generating a binary warning decision. The styles of hitting the gas/brake pedal (e.g., aggressive or gentle) are not discussed in this research. For more information about analyzing driving styles, readers are referred to \cite{xu2015establishing,mulder2011design,wang2017driving}.

\subsection{Paper Organization}
The remainder of this paper is organized as follows. Section II formulates the problem to be solved. Section III presents the GMM-HMM method for learning and inferring drivers' braking actions. Section IV describes the experiment and data collection. Section V provides the results and analysis. Section VI discusses conclusions and suggests future works.

\begin{figure}[t]
	\centering
	\includegraphics[width = \linewidth]{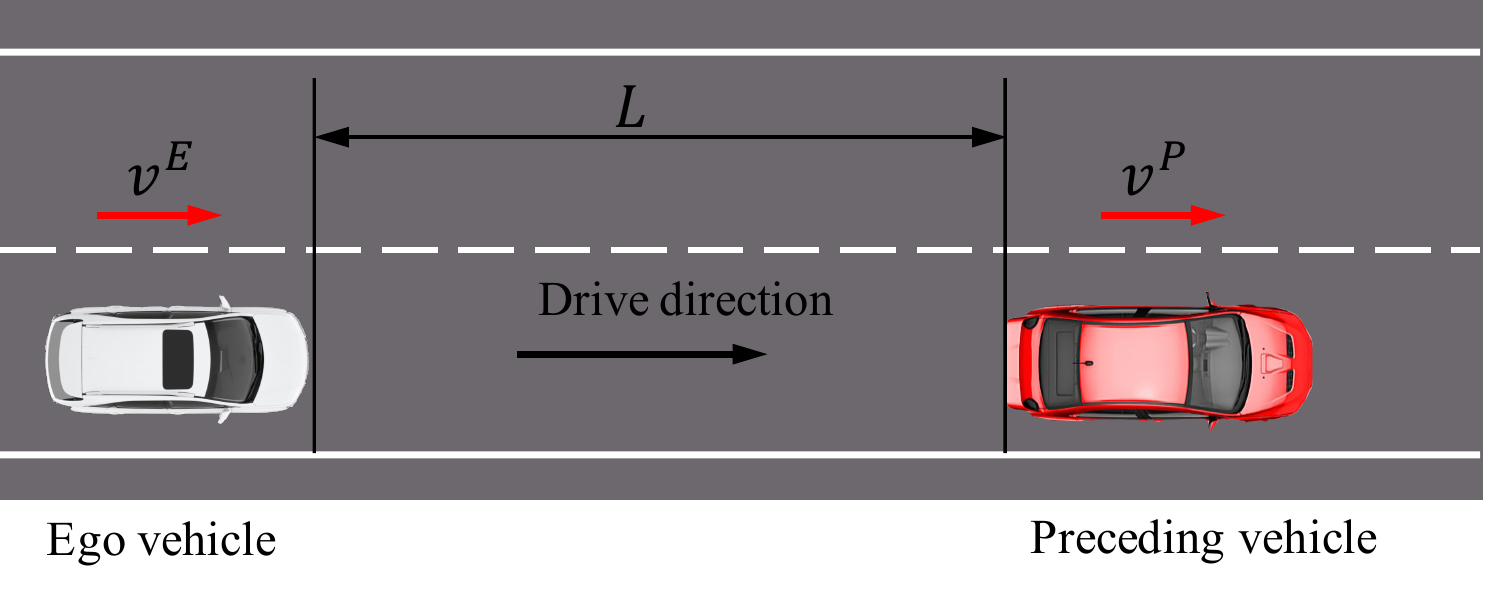}
	\caption{Car-following scenarios on a two-lane road.}
	\label{figure2}
\end{figure}

\section{Problem Formulation}
In this section, before introducing the GMM-HMM method, we first define the car-following scenario and discuss the braking behavior model we aim to build.
\subsection{Car-Following Scenarios}
Two vehicles (i.e., ``ego vehicle'' and ``preceding vehicle'') are involved in the car-following scenario, as shown in Fig. \ref{figure2}. The car-following scenario is defined using the following criteria:

\begin{itemize}
	\item We mainly infer the driver's braking action of the ego vehicle.
	\item The preceding vehicle is the vehicle located ahead in the same lane as the ego vehicle. The surrounding vehicles that are in the different lane as the ego vehicle are not considered.
	\item The relative distance, $ L $, between the ego vehicle and the preceding vehicle is less than $ 120 $ m. If $ L \geq 120$ m, the ego vehicle is in a free-following case \cite{higgs2015segmentation}.
	\item The ego vehicle is driving on roads with a small curvature, $ \rho \leq 10^{-3} \ \mathrm{m}^{-1} $. The car-following behavior on road with a big curvature is not considered. Curvature is limited to avoid the case where the roadway or the preceding vehicle in front of the ego vehicle is out of radar detecting range. 
\end{itemize}
Only longitudinal control is involved, consisting of braking and accelerating behaviors. Steering behaviors during car following are not included in this work. Interested readers can refer to \cite{guononlinear} for lateral and longitudinal control research on collision avoidance.

\subsection{Braking Behavior Model}
As discussed in the foregoing section, we aim to describe a driver's braking behavior from the perception-decision-action \cite{windridge2013characterizing} perspective. Drivers normally make braking decisions and then conduct braking actions according to their internal model and perceptions of the \textit{driving situations} they are in.  Drivers are more likely to be subject to a variety of decision-making rules because of variances in individuals' experiences and dynamic driving situations. For instance, some drivers prefer to follow the lead vehicle closely, while others prefer a greater headway. Since our goal is to predict the driver's braking actions based on information of driving situations, we will develop a model that can generate sequences which are as close as possible to what the driver would have done in the same situations, thus being able to infer this person's braking action. In what follows, the driving situations and input/output of the developed model are detailed.

\subsubsection{Explanatory Variables}
In this research, \textit{driving situation} is modeled by the states from which the driver of the ego vehicle can extract information, make decisions, and execute actions. The states are derived from the ego vehicle and the preceding vehicle. Thus, the \textit{driving situation} can be described using the following variables:

\begin{itemize}
	\item {\it Time to collision} (TTC): TTC has been widely accepted as the basic criterion for designing various types of FCW systems and car-following models \cite{wang2016development, wang2016forward, takada2014effectiveness,milanes2012fuzzy}. The basic TTC is defined as the time it would take the cars to collide at their present speed. There are many extended versions of TTC, such as Honda's TTC algorithm \cite{fujita1995radar}, which takes an additional safety margin into consideration. More extended TTC-algorithms are described in \cite{wang2016development}. In order to make computation easier, in this work, the basic TTC  at time $ t $ is obtained by:

	\begin{equation}\label{eq1}
	TTC_{t} = \frac{L_{t}}{v^{E}_{t}}
	\end{equation}
	where $ v^{E}_{t} $ is the speed of the ego vehicle and $ L_{t} $ is the relative distance between the ego and preceding vehicles at time $ t $.
	
	\item {\it Relative speed} ($ \Delta v $): Even though drivers have a relatively limited ability to perceive the absolute value of longitudinal distance or acceleration, they are good at estimating relative kinematics such as changes in relative spacing or relative speed \cite{gray1998accuracy,pariota2015linear,warren1995self}. Therefore, the relative speed between the ego vehicle and the preceding vehicle is used as a feature to describe a driver's decision to brake. It can be computed by $  \Delta v  = v^{P} - v^{E} $, where $ v^{P} $ is the speed of the preceding vehicle.
	
	\item {\it Relative distance} ($ L $):  Taieb-Maimon and Shinar \cite{taieb2001minimum} found that drivers have the ability to adjust the relative distance $ L $ by hitting brake or acceleration pedal to keep themselves at what they felt to be a `comfort' distance when following a lead car. Some drivers, for example, tend to keep a large relative distance, but some drivers do not.
	
	\item {\it Speed of the ego vehicle} ($ v^{E} $): The speed of the ego vehicle will also influence drivers' braking decision \cite{boer1999car}. Differences in psychological and physiological perceptions among drivers result in various decisions of keeping the vehicle speed at their own `comfort' level\cite{richard2012motivations}. The longitudinal speed is directly controlled by the driver's braking or accelerating actions.
\end{itemize} 
In the car-following scenario, therefore, we used four variables, including $ TTC $, $ \Delta v $, $ L $, and $ v^{E} $, to describe the driving situation perceived by the driver of the ego vehicle. 

\subsubsection{Inferring Braking Action}
Given a specific driving situation, we will infer drivers' braking action $ Br_t $ from the learned model. Note that we focus primarily on inferring if the driver will brake, thus $ Br_{t}$ can be described by a binary variable $ Br_{t} \in \{1, 0\} $, with $ 1 $ and $ 0 $ representing `brake' and `no brake', respectively. The model with the aim to infer a driver's braking action at time $ t $ based on historical data ($ Br_{1:t-1} $ and $ \boldsymbol{\xi}_{1:t-1} $) and the current observable states $ \boldsymbol{\xi}_{t} $ can be formulated as

\begin{equation}\label{Eq2}
f(Br_{t};Br_{1:t-1}, \boldsymbol{\xi}_{1:t}): \boldsymbol{\xi}_{t} \mapsto Br_{t}
\end{equation}
where $ \boldsymbol{\xi}_{t} = \{ L_{t}, v^{E}_{t}, \Delta v_{t},  TTC_{t} \}^{\top} \in \mathbb{R}^{4 \times 1}$, $ Br_{t} $ is the driver's braking decision at time $ t $, and $ f $ is the model we aim to learn. 

\section{Methodology}

\begin{figure}[t]
	\centering
	\includegraphics[width = \linewidth]{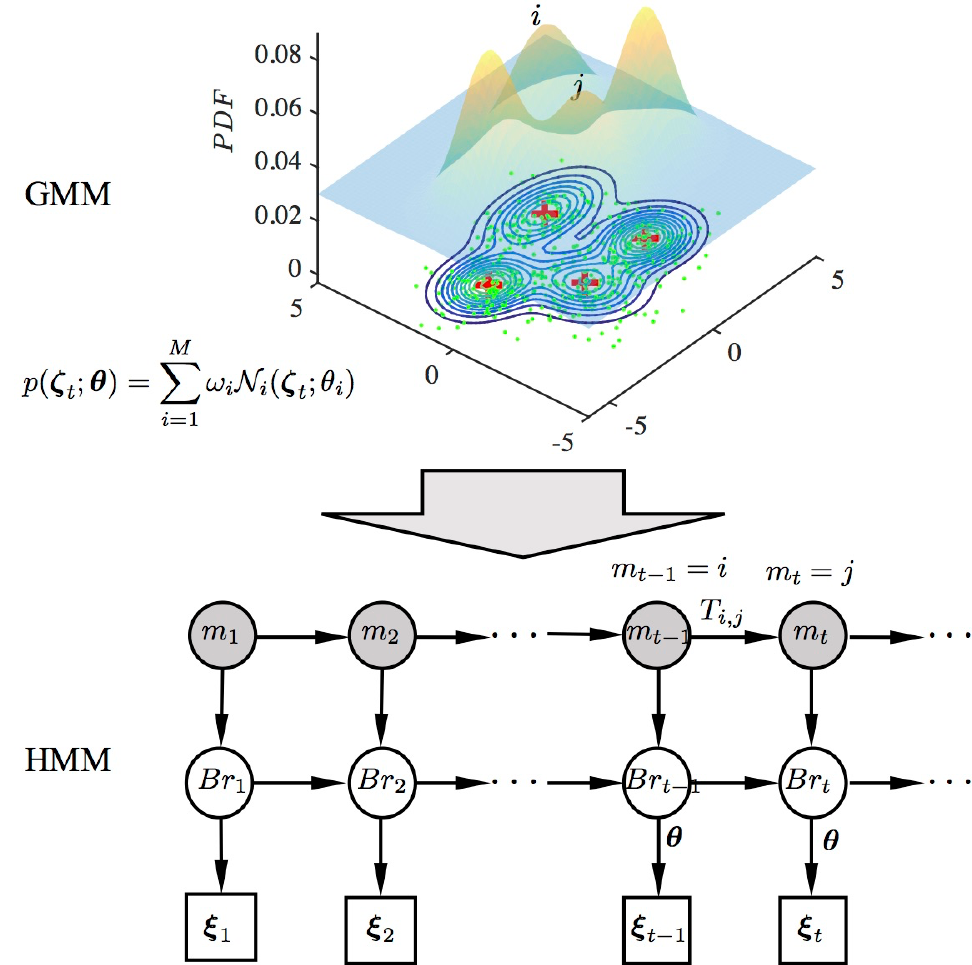}
	\caption{An illustration of the developed GMM-HMM method for inferring drivers' braking actions, where $ m_{t} $ is the hidden mode, $ \boldsymbol{\xi}_{t} $ is the observable state, $ Br_{t} $ is the unobservable state, and $ \boldsymbol{\theta} $ is the emission parameter.}
	\label{basic_idea}
\end{figure}

Most of the previous research formulates drivers' car-following behavior using fundamental equations with several physical variables\cite{pariota2015linear,jiang2015some,brackstone1999car}, but it is difficult to estimate the model parameters and describe stochastic and dynamic features of driver behaviors. Fortunately, HMM has shown a powerful ability to model and explain dynamic behavior of human driver \cite{tang2016modeling,tran2012modeling,gadepally2014framework} and thus been widely used. Based on the structure of HMM, we developed a model to describe drivers' braking behavior in car-following scenarios by combining it with GMM, as shown in Fig. \ref{basic_idea}. We selected the GMM to formulate the relationship between the state of driving situation and the braking action while keeping the stochastic feature of braking behavior. Each mixture component of GMM is treated as a hidden mode of the HMM. The GMM is chosen because it has demonstrated its powerful effectiveness in modeling other driving tasks and the stochastic features of driver behavior \cite{butakov2015personalized,lefevre2016learning,angkititrakul2011use,wang2017learning,lefevre2015driver}. The basic concepts of GMM-HMM are discussed in the following sections.

\subsection{Gaussian Mixture Model}
In order to determine the hidden mode of HMM, we define an augmented vector $ \boldsymbol{\zeta}  = [\boldsymbol{\xi}^{\top}, Br]^{\top} \in \mathbb{R}^{5 \times 1}$ to describe the relationships between driving situations and braking actions. The joint probability density function between the driving situation and the braking action is presented in the form of a multivariate Gaussian regression function:

\begin{equation}\label{eq3}
\begin{split}
p(\boldsymbol{\zeta}_{t} ; \boldsymbol{\theta})  = & \sum_{i=1}^{M} \omega_{i}\mathcal{N}_{i}(\boldsymbol{\zeta}_{t}; \theta_{i}) \\
 = & \sum_{i=1}^{M} \omega_{i} \frac{1}{(2\pi)^{d/2}|\boldsymbol{\Sigma}_{i}|^{1/2}} \\
& \times \exp\left\lbrace -\frac{1}{2}  (\boldsymbol{\zeta}_{t} - \boldsymbol{\mu}_{i})^{\top} (\boldsymbol{\Sigma}_{i})^{-1} (\boldsymbol{\zeta}_{t} - \boldsymbol{\mu}_{i}) \right\rbrace 
\end{split}
\end{equation}
with $ \boldsymbol{\theta} =\{\omega_{i}, \theta_{i}\}_{i=1}^{M} $, where $ \mathcal{N}_{i}(\boldsymbol{\zeta}_{t};\theta_{i}) $ is  the $ i $-th multivariate Gaussian distribution of dimension $ d $ (Here, $ d = 5 $); $ M \in \mathbb{N}^{+}$  is the number of Gaussian components; $ \theta_{i} = (\boldsymbol{\mu}_{i}, \boldsymbol{\Sigma}_{i})$, $ \boldsymbol{\mu}_{i} $ and $ \boldsymbol{\Sigma}_{i} $ are the mean and covariance of the $ i $-th Gaussian component; $ \omega_{i} $ is the weight of the $ i $-th mixture component and $ \sum_{i=1}^{M} \omega_{i}= 1 $.

Given a data set for a particular driver, the GMM parameter $ \boldsymbol{\theta} $ can be estimated using the maximization-likelihood (ML) method. We assume that the training data set is $ \mathcal{S}_{Train} =\{ \boldsymbol{\zeta}_{1}, \boldsymbol{\zeta}_{2}, \cdots, \boldsymbol{\zeta}_{t}, \cdots, \boldsymbol{\zeta}_{n} \} $ with time intervals $ t $ denoted by natural numbers, and the goal of the ML method is to find the parameter $ \boldsymbol{\theta} $ that maximizes the likelihood of the GMM function

\begin{equation}\label{Eq4}
\mathcal{L}(\boldsymbol{\theta}) =  \underset{\boldsymbol{\theta}}{\max} \sum_{t=1}^{n}\log (p(\boldsymbol{\zeta}_{t}; \boldsymbol{\theta})).
\end{equation}
However, the non-linearity of (\ref{Eq4}) with regard to $ \boldsymbol{\theta} $ limits to search the optimal value by directly solving (\ref{Eq4}). Fortunately, the Expectation-Maximum (EM) \cite{xuan2001algorithms} algorithm provides a possible means to get the optimal value of $ \boldsymbol{\theta} $ that maximizes $ \mathcal{L}(\boldsymbol{\theta}) $ with iteration. The EM algorithm can guarantee a monotonic increase for $ \mathcal{L} (\boldsymbol{\theta}) $ at each iterated step with the aim of finding a set of $ \boldsymbol{\theta} $ that maximize (\ref{Eq4}).  Assuming  $ \widehat{\boldsymbol{\theta}}^{k}  $ is the estimation of $ \boldsymbol{\theta} $ at iteration step $ k $, we can update $ \widehat{\boldsymbol{\theta}} $ as follows. For each iteration step, we finish the E-step and M-step by:

\begin{itemize}
	\item (\textbf{E-Step}): The {\it posterior} probability for each component $ i $ is obtained from the previous iteration $ \widehat{\boldsymbol{\theta}}^{k}  $:
	\begin{equation}\label{eq5}
	\mathrm{P}(\boldsymbol{\zeta}_{t})^{k+1}_{i} = \frac{\widehat{\omega}_{i}^{k} \mathcal{N}_{i}(\boldsymbol{\zeta}_{t}; \widehat{\boldsymbol{\mu}}^{k}_{i}, \widehat{\boldsymbol{\Sigma}}^{k}_{i})}{\sum_{j=1}^{M}\widehat{\omega}_{j}^{k} \mathcal{N}_{j}(\boldsymbol{\zeta}_{t}; \widehat{\boldsymbol{\mu}}^{k}_{j}, \widehat{\boldsymbol{\Sigma}}^{k}_{j})}.
	\end{equation}
	
	\item (\textbf{M-Step}): Then, update the model parameter $ \widehat{\boldsymbol{\theta}} $ at step $ k+1 $ by
	\begin{equation}\label{eq6}
	\widehat{\omega}^{k+1}_{i} = \frac{1}{n} \sum_{t=1}^{n}\mathrm{P}(\boldsymbol{\zeta}_{t})^{k+1}_{i},
	\end{equation}
	
	\begin{equation}\label{eq7}
	\widehat{\boldsymbol{\mu}}^{k+1}_{i} = \frac{\sum_{t=1}^{n}\mathrm{P}(\boldsymbol{\zeta}_{t})^{k+1}_{i} \boldsymbol{\zeta}_{t}}{\sum_{t=1}^{n}\mathrm{P}(\boldsymbol{\zeta}_{t})^{k+1}_{i} },
	\end{equation}
	
	\begin{equation}\label{eq8}
	\widehat{\boldsymbol{\Sigma}}^{k+1}_{i} = \frac{\sum_{t=1}^{n} \mathrm{P}(\boldsymbol{\zeta}_{t})^{k+1}_{i} (\boldsymbol{\zeta}_{t} - \widehat{\boldsymbol{\mu}}^{k+1}_{i}) (\boldsymbol{\zeta}_{t} - \widehat{\boldsymbol{\mu}}^{k+1}_{i})^{\top}}{\sum_{t=1}^{n}\mathrm{P}(\boldsymbol{\zeta}_{t})^{k+1}_{i}}.
	\end{equation}
	\item Update the value of log-likelihood function $ \mathcal{L}(\widehat{\boldsymbol{\theta}}^{k+1}) $ by
	
	\begin{equation}\label{eq9}
	\mathcal{L}(\widehat{\boldsymbol{\theta}}^{k+1}) = \sum_{t=1}^{n} \log (p(\boldsymbol{\zeta}_{t}; \widehat{\boldsymbol{\theta}}^{k+1}))
	\end{equation}
	where $  \widehat{\boldsymbol{\theta}}^{k+1} = (\widehat{\omega}^{k+1}_{i}, \widehat{\boldsymbol{\mu}}^{k+1}_{i}, \widehat{\boldsymbol{\Sigma}}^{k+1}_{i})_{i=1}^{M}$.
\end{itemize}
 Update (\ref{eq5}) -- (\ref{eq9}) until  the log-likelihood converges, i.e., $ \mathcal{L}(\widehat{\boldsymbol{\theta}}^{k+1}) - \mathcal{L}(\widehat{\boldsymbol{\theta}}^{k}) < \epsilon  $. In this work, we set $ \epsilon = 10^{-10} $.  The number $ M $ of Gaussian components is determined using the Bayesian information criterion (BIC) \cite{calinon2010learning}. We can thus get the optimally estimated parameter of GMM, denoted by $ \boldsymbol{\theta}^{\ast} $
 
 \begin{equation}\label{eq10}
\boldsymbol{\theta}^{\ast} = \arg  \underset{\boldsymbol{\theta}}{\max} \  \sum_{t=1}^{n} \log (p(\boldsymbol{\zeta}; \boldsymbol{\theta})).
 \end{equation}

\subsection{Hidden Markov Model}
After learning the GMM parameters, each mixture component of GMM is treated as a hidden mode of HMM, thus obtaining an HMM with $ M $ hidden states. Based on the trained GMM consisting of $ M $ multivariate Gaussians, we obtain a corresponding mode $ m_{t} \in \{ 1,2,\cdots, M \} $ given an observed data $ \boldsymbol{\zeta}_{t} $ at time $ t $, as shown in Fig. \ref{basic_idea}. Our goal is to infer drivers' braking actions from the defined driving situation, so we define the following variables for the HMM:

\begin{itemize}
	\item \textit{Hidden mode}: we define $ \mathcal{M}_{t} \in \{ 1,2,\cdots, M \}$ as the hidden mode at time $ t $, with $ M $ the number of possible hidden modes.
	\item \textit{Observable state}: $ \mathcal{O}_{t} = \boldsymbol{\xi}_{t} = [L_{t}, v^{E}_{t}, \Delta v_{t}, TTC_{t}]$ is the observable state at time $ t $.
	\item \textit{Unobservable state}: $ \mathcal{U}_{t} = Br_{t}$ is the hidden state we need to infer from the observable state $ \mathcal{O}_{t} $ at time $ t $.
	\item \textit{Transfer matrix}: $ \mathcal{T} = \{ T_{i,j} \}_{i,j}^{M} \in \mathbb{R}^{M\times M}$ is the transfer probability matrix. $ T_{i,j} $ is the transfer probability from mode $ i $ to mode $ j $.
	\item \textit{Emission parameters}: $ \mathcal{E} = \{ \boldsymbol{\mu}_{i}, \boldsymbol{\Sigma}_{i} \}_{i = 1}^{M} $ are the emission parameters from the hidden states to the observable states.
\end{itemize}
The transfer probability $ T_{i,j} $ can be estimated from the training data set $ \mathcal{S}_{Train} $ (see Appendix A). Thus, the HMM model can be represented by $ \{ \mathcal{M}_{t},  \mathcal{O}_{t}, \mathcal{U}_{t}, \mathcal{T}, \mathcal{E} \} $ based on the learned GMM. In GMM-HMM, the joint distribution between the hidden modes and the states consisting of observable and unobservable states is formulated by 

\begin{equation}
\begin{split}
p(\mathcal{M}_{0:t},\mathcal{O}_{1:t},\mathcal{U}_{1:t}) = & p(\mathcal{M}_{0})   \times \\
 & \prod_{k=1}^{t}\left[ p(\mathcal{M}_{k}|\mathcal{M}_{k-1}) \cdot p(\mathcal{O}_{k},\mathcal{U}_{k}|\mathcal{M}_{k})\right] \\
= & p(\mathcal{M}_{0}) \prod_{k=1}^{t} \left[ T_{k-1,k}\cdot p(\mathcal{O}_{k},\mathcal{U}_{k}|\mathcal{M}_{k}) \right].
\end{split}
\end{equation} 

The braking action at time $ t $ is inferred from the consecutive values of the driving situation using GMM, i.e., $ \widehat{Br}_{t} $ is obtained as the conditional expectation of $ Br_{t} $ given the sequence $ \boldsymbol{\xi}_{1:t} $\cite{lefevre2016learning},

\begin{equation}\label{eq11}
\begin{split}
\widehat{Br}_{t} = & E(Br_{t};\boldsymbol{\xi}_{1:t}) \\
= & \sum_{i=1}^{M} \alpha_{i,t} \left(  \boldsymbol{\mu}^{Br}_{i} + \boldsymbol{\Sigma}_{i}^{Br,\boldsymbol{\xi} } (\boldsymbol{\Sigma}_{i}^{\boldsymbol{\xi},\boldsymbol{\xi}})^{-1}  (\boldsymbol{\xi}_{t} - \boldsymbol{\mu}_{i}^{\boldsymbol{\xi}}) \right) 
\end{split}
\end{equation}
where 

\begin{equation}\label{eq12}
\boldsymbol{\mu}_{i} = 
\begin{bmatrix}
\boldsymbol{\mu}_{i}^{\boldsymbol{\xi}} \\
\boldsymbol{\mu}_{i}^{Br} \\
\end{bmatrix}
\end{equation}

\begin{equation}\label{eq13}
\boldsymbol{\Sigma}_{i} = 
\begin{bmatrix}
\boldsymbol{\Sigma}_{i}^{\boldsymbol{\xi}, \boldsymbol{\xi}} & \boldsymbol{\Sigma}_{i}^{\boldsymbol{\xi}, Br}\\
\boldsymbol{\Sigma}_{i}^{Br, \boldsymbol{\xi}} & \boldsymbol{\Sigma}_{i}^{Br, Br}\\
\end{bmatrix},
\end{equation}
and $ \alpha_{i,t} $ is the HMM forward variable, calculated by 

\begin{equation}\label{eq14}
\alpha_{i,t} = \frac{\left(  \sum_{j=1}^{M} \alpha_{j,t-1}T_{j,i} \right)  \mathcal{N}_{i}(\boldsymbol{\xi}_{t}; \boldsymbol{\mu}_{i}^{\boldsymbol{\xi}},\boldsymbol{\Sigma}_{i}^{\boldsymbol{\xi},\boldsymbol{\xi} })}{\sum_{k=1}^{M} \left[ \left(  \sum_{j=1}^{M} \alpha_{j,t-1}T_{j,k} \right)  \mathcal{N}_{k}(\boldsymbol{\xi}_{t}; \boldsymbol{\mu}_{k}^{\boldsymbol{\xi}},\boldsymbol{\Sigma}_{k}^{\boldsymbol{\xi},\boldsymbol{\xi} }) \right] }.
\end{equation}
The initial value with $ t=1 $ is given by
\begin{equation*}
\alpha_{i,1} = \frac{\omega_{i}\mathcal{N}_{i}(\boldsymbol{\xi}_{1}; \boldsymbol{\mu}_{i}, \boldsymbol{\Sigma}_{i}^{\boldsymbol{\xi},\boldsymbol{\xi}})}{\sum_{k=1}^{M}\omega_{k}\mathcal{N}_{k}(\boldsymbol{\xi}_{1}; \boldsymbol{\mu}_{k}, \boldsymbol{\Sigma}_{k}^{\boldsymbol{\xi},\boldsymbol{\xi}})}.
\end{equation*}

Based on the above steps, drivers' braking actions $ Br_{t} $ at time $ t $ can be inferred from historic samples $ \boldsymbol{\zeta}_{1:t-1} $ and its present driving conditions $ \boldsymbol{\xi}_{t} $ (i.e., observable states, $ \mathcal{O}_{t} $).

\begin{figure}[t]
	\centering
	\includegraphics[width = \linewidth]{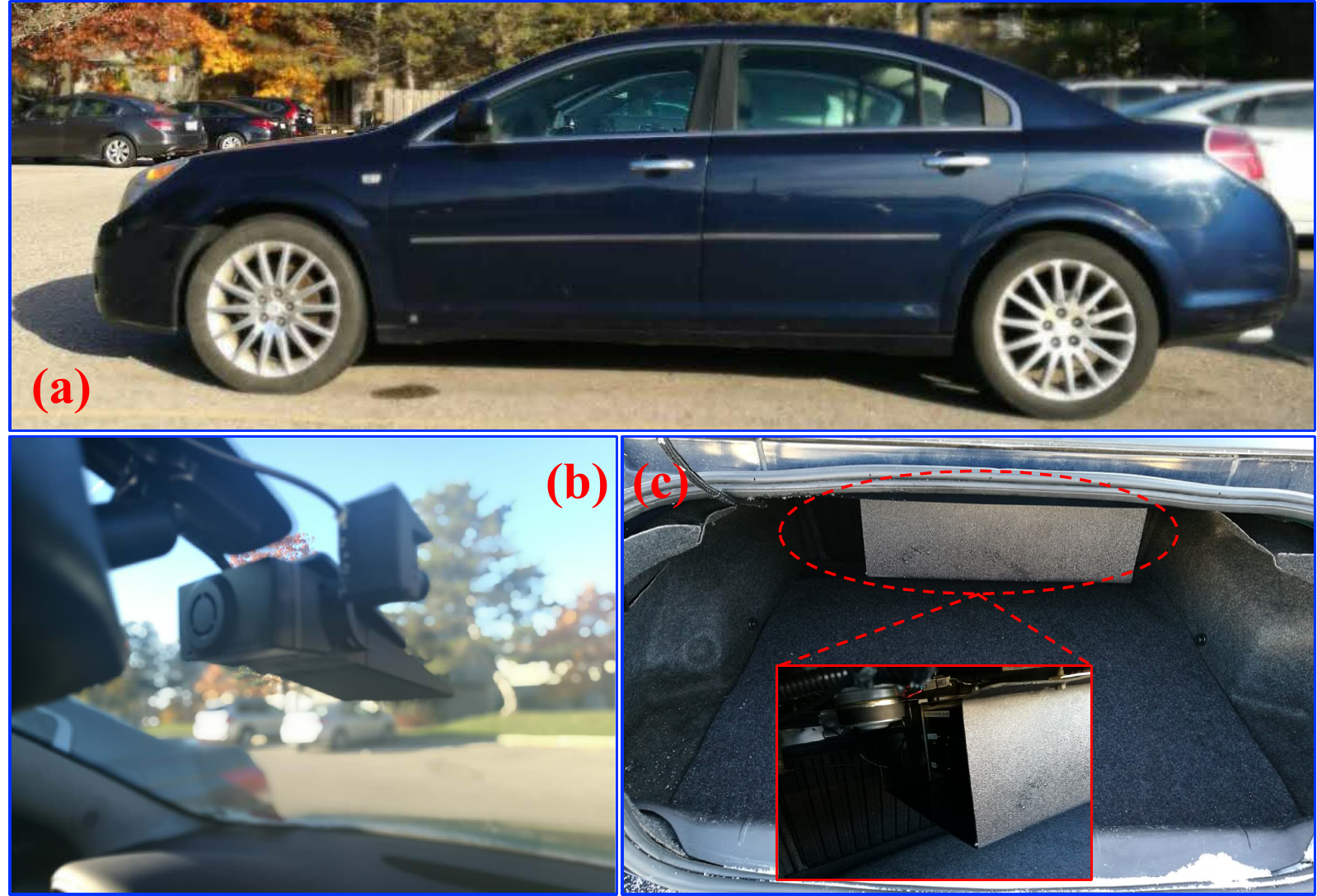}
	\caption{One of the experimental vehicles with data-collection equipment. (a) An experimental vehicle; (b) Mobileye; (c) Data acquisition systems.}
	\label{exps}
\end{figure}

\begin{figure*}[t]
	\centering
	\includegraphics[width = 0.85\textwidth]{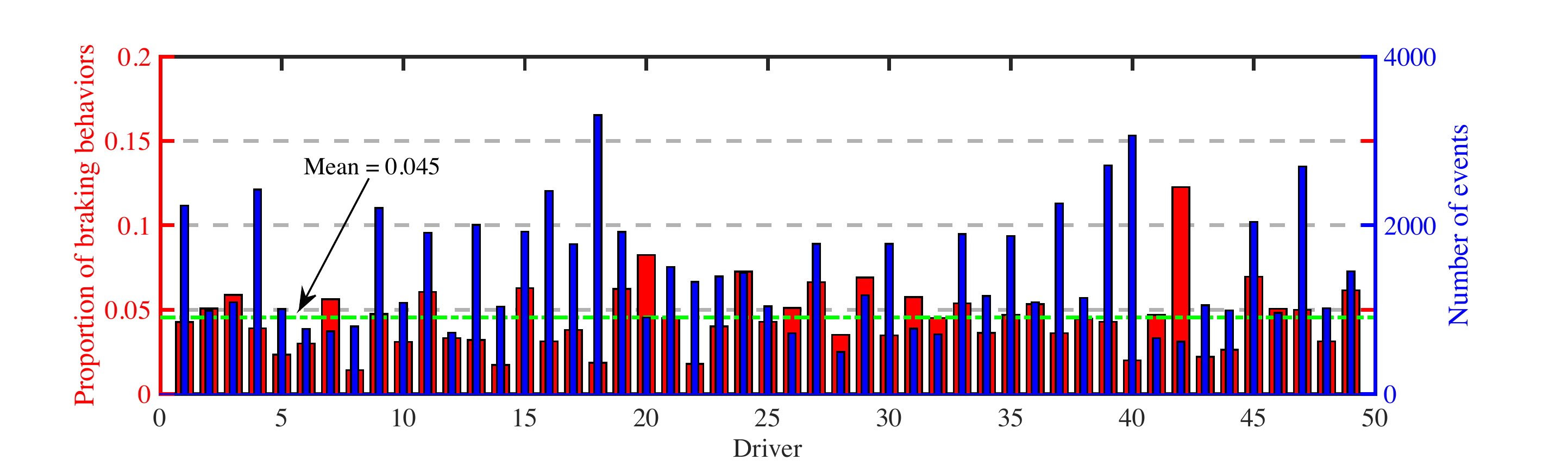}
	\caption{The proportion of braking behaviors in the entire driving data and the number of car-following events for different drivers.}
	\label{BrakeProp}
\end{figure*}

\section{Experiment and Data Collection}
We used naturalistic driving data collected from the University of Michigan Safety Pilot Model Deployment (SPMD) program \cite{bezzina2014safety}. It recorded the naturalistic driving of 2,842 equipped vehicles in Ann Arbor, Michigan, for over three years. As of April 2016, 34.9 million miles of driving data were logged, making the SPMD database one of the largest public N-FOT ones ever. In the SPMD database, 98 sedans were equipped with data acquisition systems (DASs) and MobilEye (Fig. \ref{exps}). The vehicle was pre-equipped from the factory with sensors to measure the speed, lane marks, relative speed and distance between the ego vehicle and the preceding vehicle, and road curvature. The data of relative speed, relative range, and road curvature were collected from the MobileEye. The ego vehicle speed, brake pedal position, and throttle opening were recorded from the CAN-bus of each vehicle.

\subsection{Driver Participants and Data Record}
In this work, 56 drivers were included. They all hold a valid driving license. Each driver performed casual daily trips without any restrictions or requirements on their trips. The drivers were not restricted to a particular route and were given no restrictions, including the duration of routes. While driving, the on-board PC recorded driving data at a frequency of 10 Hz. The DASs were not shown to the drivers and the process of recording data were hidden from the drivers, guaranteeing that the drivers were not disturbed by the DASs and that the recorded data were naturalistic. While driving, many kinds of driving behaviors, such as lane changing, overtaking, and car-following, were recorded.

\subsection{Data Extraction}
In order to extract the car-following data from the database, we defined the car-following {\it event} as follows:

\begin{itemize}
	\item The relative distance $ L $ between the ego vehicle and preceding vehicle should be less than 120 m \cite{higgs2015segmentation}. If the relative distance was $ L >120 $ m, then we ended the event.
	\item The ego vehicle and the preceding vehicle should be in the same lane. If the left or right turn light was on, which indicates the driver would make a lane change or overtaking behavior, then the event was ended.
	\item If one other vehicle was merging between the ego vehicle and the preceding vehicle, the car-following event was ended. 
	\item When vehicle speed was less than 5 m/s (i.e., 18 km/h), the event ended. In addition, if the relative range was less than 10 m, the event ended, ensuring that no Stop-\&-Go case was included.
	\item The duration for a singular car-following event should be larger than 50 s. 
	\item Any driver with less than 500 car-following events was eliminated, which was able to guarantee the collected data were enough to capture the underlying driving styles\cite{wang2017much}.
\end{itemize}
Based on the above criteria, 49 drivers were selected out from 56 drivers. The number of car-following events for each driver was about 1,480 and each \textit{event} lasted about 86.93 seconds on average. The extracted data consisted of the variables as follows:
%
%

\begin{enumerate}
	\item Vehicle speed, $ v^{E} \in [5, 45] $ [m/s];
	\item Relative speed, $ \Delta v $;
	\item Relative range, $ L \in[10, 120] $ [m];
	\item Time to collision, $ TTC $, calculated from (1)
	\item Throttle opening, $ Th \in [0,100] $;
	\item Braking action, $ Br = 1\ \mathrm{or} \ 0 $. 
\end{enumerate}
The braking actions $ Br $ were recorded with the rules: $ Br = 1$ when the driver put his/her foot on the brake pedal; otherwise $ Br = 0 $ when the driver removed his/her foot from the brake pedal.

\subsection{General Data Analysis}

Overall, we collected 72,166 car-following events from the 49 drivers. Fig. \ref{BrakeProp} shows the number of car-following events and the percentage of data with braking actions in all events of each driver. We found that the percentage of those pushing the brake pedal during the car-following task was about 4.5\% on average, as shown by the green dashed line in Fig. \ref{BrakeProp}.  Some drivers tended to prefer frequent braking actions, (e.g., driver \#42 with 12.27\% braking actions); others tended to brake less often (e.g., driver \#8 with 1.43\% braking actions). When we preprocessed data, the driver's {\it movement time} -- the time to lift the foot off the accelerator pedal, move it laterally to the brake pedal, or vice versa -- was not considered because the performance time was very small (about 0.15 -- 0.30 seconds \cite{davies1969preliminary}), compared to the total driving time and the time of keeping their foot on the brake pedal. We also found that drivers would not always put their foot on the brake pedal and gas pedal, but the duration was very small, compared to the total car-following time (with a percentage of about $ 1.873\times10^{-4} $ in our experiments).  For each driver, in total, we obtained 95.6 minutes of driving data on average with keeping their foot on the brake pedal. 

\begin{figure}[t]
	\centering
	\includegraphics[width = \linewidth]{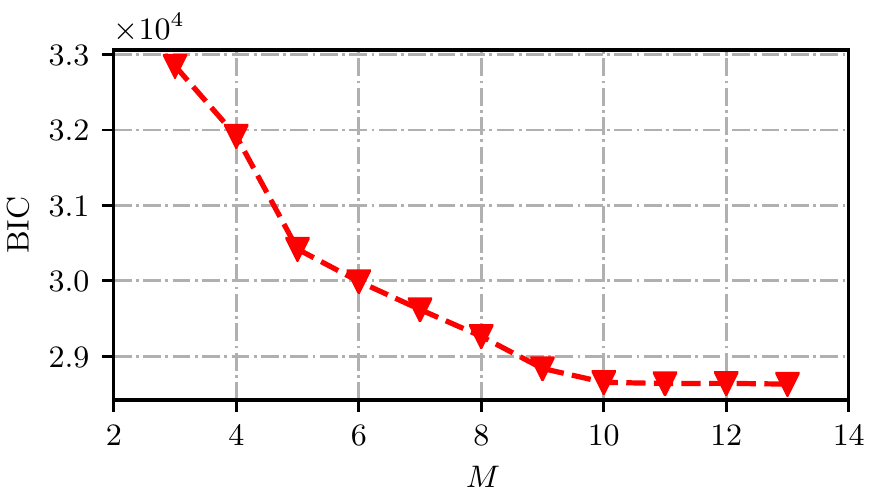}
	\caption{An example of test results of BIC with different numbers of GMM components for driver \#4, with total 2,233 events.}
	\label{BIC}
\end{figure}

\subsection{Methods for Comparison}
In this paper, the driver participants' braking actions were recorded as a binary variable. Therefore, the task of inferring whether the driver will brake when following a preceding car can be interpreted as a supervised classification problem, thus providing an opportunity for ones to utilize the standard methods capable of dealing with the binary classification. The support vector machine (SVM) and its extensions have shown advantages in recognizing driver intents \cite{kumar2013learning} and driving styles\cite{wang2017driving,wang2016rapid}. Therefore, we selected two SVM-based approaches to make comparisons, one was the basic SVM method \cite{kumar2013learning} and the other one was the SVM-BF method\cite{aoude2012driver} that combines SVM and Bayesian filtering. 

\subsubsection{SVM} Given a binary labeled training data $ \{ \boldsymbol{\xi}_{i}, Br_{i} \} $, where $ Br_{i} $ is the label specified with 1 or 0 of the observation $ \boldsymbol{\xi}_{i} $, a new test data $ \boldsymbol{\xi} $ is classified into one class ($ Br = 1 $) or the other ($ Br = 0 $) based on a trained SVM classifier. Based on outputs of SVM, the drivers' braking actions can thus be directly inferred.

\subsubsection{SVM-BF} In the SVM-BF method\cite{aoude2012driver}, the output of the SVM block is then fed into a Bayesian filtering (BF) block, which provides an additional logic before making a final classification. The BF component transfers the output of the SVM block into a probability that the driver takes a braking action with present driving conditions $ \boldsymbol{\xi} $, i.e., $ p(Br = 1|\boldsymbol{\xi}) \in [0,1] $. Based on $ p(Br = 1|\boldsymbol{\xi}) $, the SVM-BF algorithm then computes the the final classification with the threshold  specified value $ \gamma_{Br} \in [0,1] $. The driver is inferred to take a braking action if $ p(Br = 1|\boldsymbol{\xi}) > \gamma_{Br} $; otherwise it is treated as no brake. In this paper, we preset the threshold to $ \gamma_{Br} = 0.9 $.

We developed the SVM and SVM-BF models based on the classification function \texttt{svm.SVC} in \texttt{scikit-learning} tools (\url{http://scikit-learn.org/stable/}) to train and test these models. Labeled datasets used to infer whether the driver will brake are not linearly separable cases because of the inherent uncertainty and nonlinearity of driver behaviors. Therefore, a nonlinear kernel, a Gaussian kernel, was selected for both of SVM and SVM-BF according to \cite{ben2010user,aoude2012driver} in this work. In order to avoid the overfitting issues, a Cross-Validation approach was also used and detailed as follows. 

\subsection{Training and Test Procedures}
\subsubsection{BIC}
The number of GMM components $ M $ was determined according to the BIC\cite{calinon2010learning}. However, the amount of driving data is various among drivers (Fig. \ref{BrakeProp}) and thus influences the BIC values. For example, a smaller amount of driving data tends to obtain a smaller optimal value of $ M $, e.g., 621 car-following events result in $ M = 9 $; on the other hand, a larger amount of driving data tends to obtain a larger optimal value of $ M $, e.g., 3,307 car-following events result in $ M = 13 $. Fig. \ref{BIC} shows an example of driving data with 2,223 car-following events, in which the BIC reaches the \textit{elbow point} when $ M = 10 $. In order to make a trade-off between the cost of computing time and the accuracy of the learned model, we selected $ M = 10 $ because a smaller value of $ M $ would reduce the fitting accuracy of the model while a larger value of $ M $ would increase the computation cost without a significant improvement in accuracy. 

\subsubsection{Cross-Validation} 
The $ \kappa $-fold \textit{Cross-Validation} (CV) method \cite{godino2006dimensionality} was used to avoid over-fitting issues and assess the model performance. For SVM and SVM-BF, the Gaussian kernel width in the \texttt{svm.SVC} was determined by parameters $ C $ and $ \gamma $, and we finally set $ C = 1.0 $ and $ \gamma = 0.01 $ according to the CV results. To do CV, for each driver, the driving data set was evenly divided into $ \kappa $ subsets, also called {\it folds}. We utilized $ \kappa -1$ folds to learn the model parameter and the left-out fold was used to assess the prediction performance, which is called {\it leave-one-out cross-validation} (LOO-CV). In this work, we selected $ \kappa=10 $. The CV method ensures that data used for learning the model parameters is disjoint from the data used for assessment. 

\subsubsection{Training and Test Procedure} For a single participant, one of ten folds was used to test, and the remaining folds were used to train, thus cyclically obtaining 10 different test results. The model performance was recorded and assessed using the average value of the 10 CV results. While training the GMM, the initial value of the GMM (i.e., the initial center of the GMM component) was determined by directly using a $ K $-means clustering method, where $ K $ was equal to the number of GMM components (i.e., $ K=M=10 $). We operated the $ K $-means clustering method for five times on the training data and selected one of the clustering results as an initial value that could maximize the likelihood function (4). 

\section{Experimental Results and Analysis}
\subsection{Performance Metrics}
In this work, our goal is to answer the question ``\textit{Will the driver brake when following a preceding car?}'', which means that the performance evaluation can be achieved using assessment approaches for binary classifications. The best performance is that the developed method can {\it correctly} infer drivers' braking/no braking actions and also can obtain a higher degree of {\it accuracy}. To better understand the meaning of {\it accuracy}, we define the following concepts based on a statistical method  in \cite{altman1994diagnostic}, as shown in Table \ref{Table_ind}:

\begin{itemize}
	\item \textit{True Positive} (TP): A TP test result is one in which inferring a braking action will occur when the driver brakes.
	\item \textit{True Negative} (TN): A TN test result is one in which inferring a braking action will not occur when the driver does not brake.
	\item \textit{False Positive} (FP):  A FP test result is one in which inferring a braking action will occur but the driver does not brake.
	\item \textit{False Negative}  (FN): A FN test result is one in which inferring a braking action will not occur when the driver brakes.
\end{itemize}
Obviously, we tend to prefer the results with a larger proportion of TP and TN. More specifically, we  define {\it accuracy}, {\it sensitivity}, and {\it specificity} as follows:
\begin{table}[t]
	\centering
	\caption{Basic Concepts of Statistical Measures.}
	\label{Table_ind}
	\begin{tabular}{c|c|c|c}
		\hline
		\hline
		\multicolumn{2}{l|}{} & \multicolumn{2}{c}{Inferring}            \\ \hline
		\multirow{3}{*}{\rotatebox[origin=c]{90}{Real}}  & \multicolumn{1}{l|}{} & Brake & No Brake \\ \cline{2-4} 
		& Positive              & TP    & FN       \\ \cline{2-4} 
		& Negative              & FP    & TN       \\ 
		\hline
		\hline
	\end{tabular}
\end{table}

\subsubsection{Accuracy} Accuracy is the commonly used measure for assessing the classification performance. In this paper, the problem of inferring whether the driver will bake was treated as a classifier problem as aforementioned. Therefore, the accuracy can be computed by 

\begin{equation}\label{eq15}
\eta_{\mathrm{acc}} = \frac{N_{TP} + N_{TN}}{N_{TP} + N_{FP} + N_{FN} + N_{TN}}
\end{equation}
where $ N_{(\ast)}$ is the number of $ (\ast) $ appearing, with $ (\ast) \in \{ TN, TP, FP, FN \} $. The accuracy metric $ \eta_{\mathrm{acc}} $ can show the comprehensive performance, with a higher value of accuracy indicating good performance.
\subsubsection{Sensitivity} Sensitivity represents the ability to correctly infer drivers' braking actions. The sensitivity is calculated by

\begin{equation}\label{eq16}
\eta_{\mathrm{sen}} = \frac{N_{TP}}{N_{TP} + N_{FN}}.
\end{equation}
A larger value of $ \eta_{sen} $ indicates that the method is more possible for correctly inferring drivers' braking actions.

\subsubsection{Specificity} Specificity is related to the ability to correctly infer that the driver does not brake. The specificity is computed by

\begin{equation}\label{eq17}
\eta_{\mathrm{spe}} = \frac{N_{TN}}{N_{TN} + N_{FP}}.
\end{equation}
Note that a larger value of $ \eta_{\mathrm{spe}} $ means that the method has a greater ability to infer the driver's `no' braking action.

\begin{figure}[t]
	\centering
	\subfloat[]{\includegraphics[width = \linewidth]{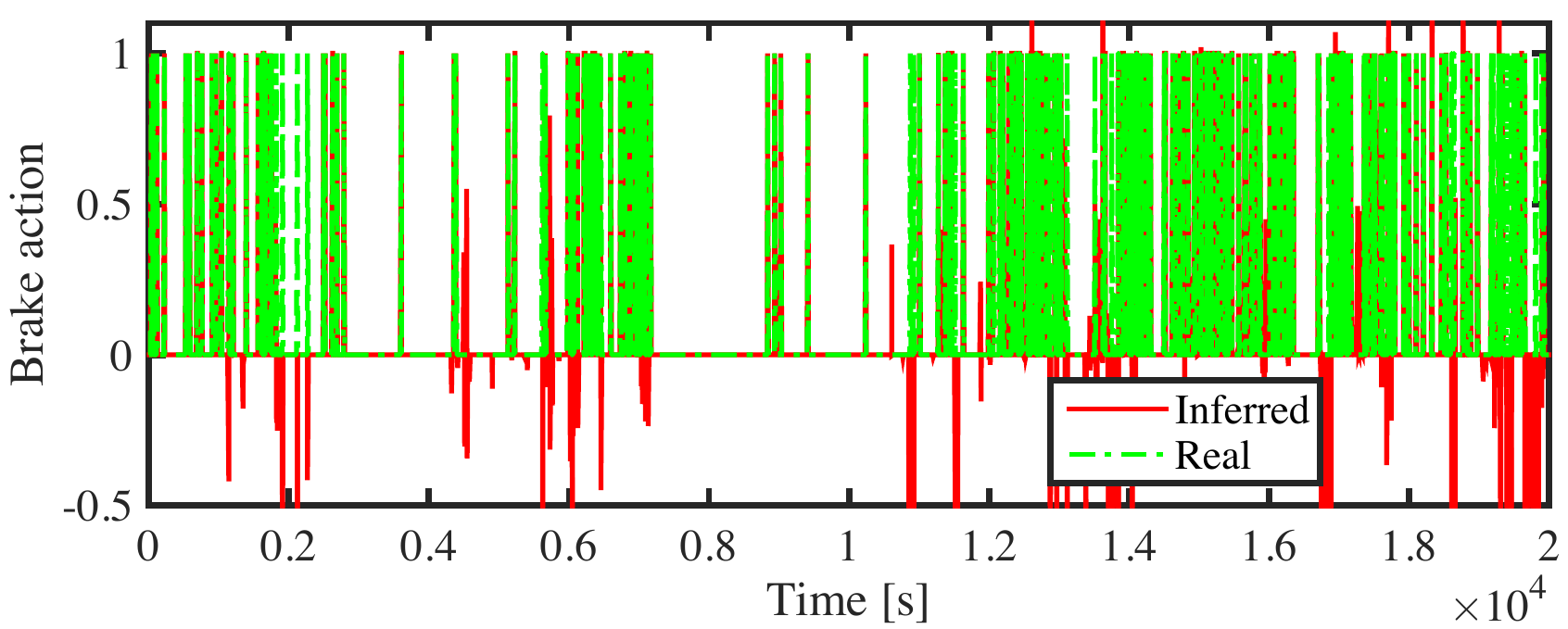}}\\
	\subfloat[]{\includegraphics[width = \linewidth]{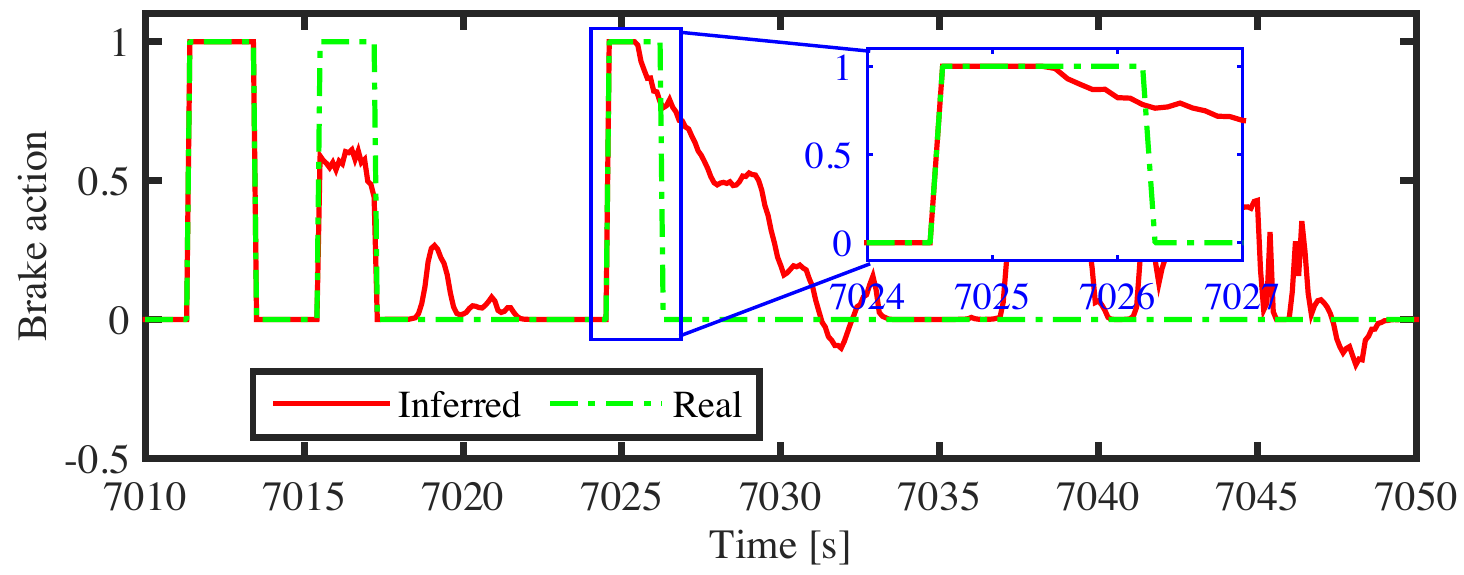}}\\
	\subfloat[]{\includegraphics[width = \linewidth]{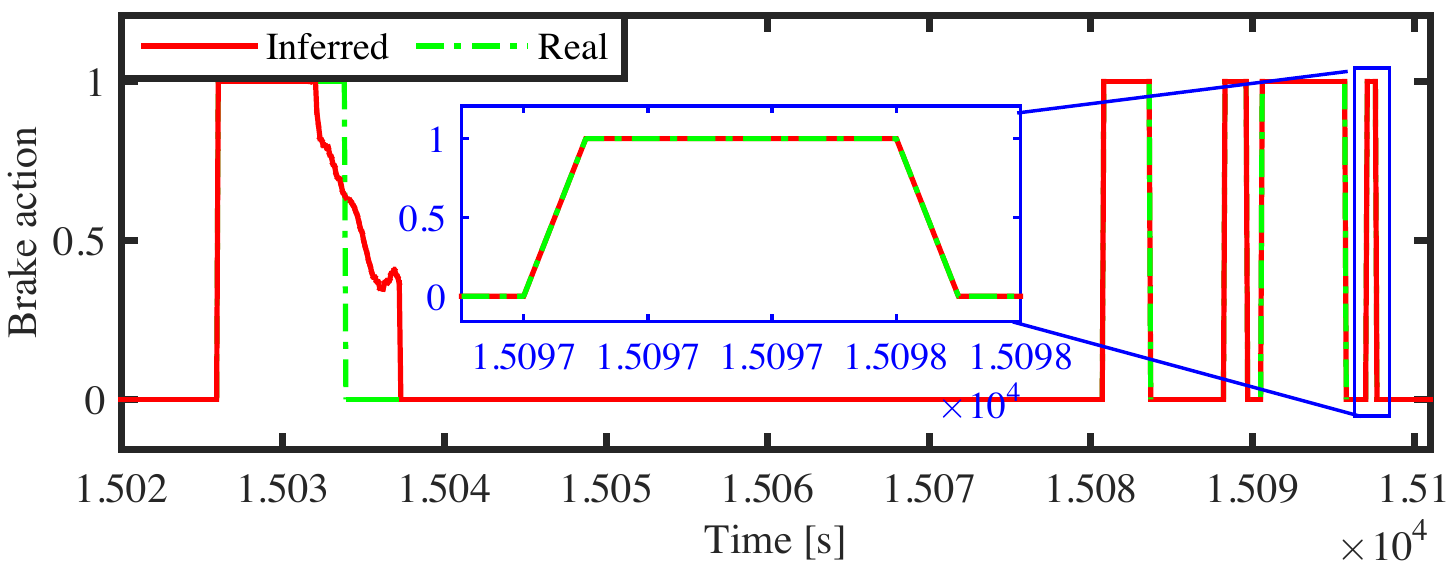}}
	\caption{Example of the inferred results using the GMM-HMM method for driver \#4 during $ 2\times 10^{4} $ seconds, including about 233 events, with $ Br^{\mathrm{c}} = 0.9 $, $ \eta_{\mathrm{acc}} = 93.87\% $, $ \eta_{\mathrm{sen}}  = 83.76\%$, and $ \eta_{\mathrm{spe}} = 98.27\% $. (a)  One of the test results for driver \#4; (b) an unreliable inferred result; and (c) a good inferred result with local data.} 
	\label{Example_res}
\end{figure}

Based on the above-mentioned definitions, we assess the performance of GMM-HMM using the three metrics, consisting of accuracy, $ \eta_{\mathrm{acc}} $, sensitivity, $ \eta_{\mathrm{sen}} $, and specificity, $ \eta_{\mathrm{spe}} $.

\subsection{Results Decoding} To compute the performance metrics (\ref{eq15}) -- (\ref{eq17}), we decode the outputs of GMM-HMM by following rules:

\begin{table}[t]
	\centering
	\caption{Performance Metrics (Mean (\%) $ \pm $ Standard Deviation) of Ten Cross-Validation Results Regarding $ \eta_{\mathrm{acc}} $, $ \eta_{\mathrm{sen}} $ and $ \eta_{\mathrm{spe}} $ for All Drivers Using the GMM-HMM Method with $ Br^{\mathrm{c}} = 0.9$ }
	\label{Table_res}
	\begin{tabular}{c|c|c|c}
		\hline
		\hline
		Driver & $ \eta_{\mathrm{acc}} $  & $ \eta_{\mathrm{sen}} $ & $ \eta_{\mathrm{spe}} $ \\
		\hline
		1       & $ 95.90 \pm 0.0125 $ & $ 86.48 \pm 0.0515 $ & $ 97.31 \pm 0.0078 $\\
		2       &  $ 90.61 \pm 0.0192 $ &  $ 87.90 \pm 0.0445 $ & $ 97.23 \pm 0.0059 $ \\
		3       & $ 93.62 \pm 0.0172$  & $ 93.99 \pm 0.0617 $ & $ 98.12 \pm 0.0162 $\\
		4       & $ 91.49 \pm 0.0191 $ & $ 89.30 \pm 0.0424 $ & $ 97.28 \pm 0.0068 $\\
		5      &  $ 94.72 \pm 0.0107 $ &  $ 81.02 \pm 0.0734 $ & $ 98.86 \pm 0.0057 $ \\
		6       &  $ 93.80 \pm 0.0232 $ &  $ 88.88 \pm 0.0633 $ & $ 98.14 \pm 0.0054 $ \\
		7       &  $ 86.56 \pm 0.1031 $ &  $ 84.85 \pm 0.0807 $ & $ 97.12 \pm 0.0077 $ \\
		8       &  $ 89.62 \pm 0.0214 $ &  $ 78.41 \pm 0.0206 $ & \colorbox{red}{$ 99.54 \pm 0.0021 $} \\
		9       & $ 91.94 \pm 0.0216 $ & \colorbox{red}{$ 95.41 \pm 0.0341 $} & $ 96.53\pm 0.0095 $\\
		10       & $ 93.61 \pm 0.0124 $ & $ 90.08 \pm 0.0736 $ & $ 98.52 \pm 0.0054 $\\
		11       & $ 89.74\pm 0.0201 $ & $ 91.95 \pm 0.0461 $ & $ 96.98 \pm 0.0072 $\\
		12       & $ 91.55 \pm 0.0261 $ & $ 88.19 \pm 0.0793 $ & $ 98.31 \pm 0.0080 $\\
		13       & $ 89.69 \pm 0.0280 $ & $ 78.26 \pm 0.0495 $ & $ 98.86 \pm 0.0038 $\\
		14       & $ 95.46 \pm 0.0176 $ & $ 84.72 \pm 0.0733 $ & $ 98.90 \pm 0.0049 $\\
		15       & $ 94.57 \pm 0.0149 $ & $ 94.44 \pm 0.0444 $ & $ 94.40 \pm 0.0138 $\\
		16       & $ 93.96 \pm 0.0094 $ & $ 73.70 \pm 0.0508$ & $ 97.63 \pm 0.0057 $\\
		17       & $ 93.78 \pm 0.0505 $ & $ 89.82 \pm 0.0234 $ & $  97.89 \pm 0.0042$\\
		18       & $ 93.27 \pm 0.0553 $ & $ 89.38 \pm 0.0984 $ & $ 99.22 \pm 0.0038 $\\
		19       & $ 93.31 \pm 0.0134 $ & $ 85.93 \pm 0.1259 $ & $ 96.48 \pm 0.0121 $\\
		20       & $ 73.10 \pm 0.0383 $ & $ 75.68 \pm 0.0585 $ & $ 94.83 \pm 0.0245 $\\
		21       & $ 94.62 \pm 0.0085 $ & $ 88.94 \pm 0.0496 $ & $ 96.69 \pm 0.0041 $\\
		22       & $ 94.50 \pm 0.0336 $ & $ 91.29 \pm 0.0594 $ & $ 99.06 \pm 0.0077 $\\
		23      & $ 82.50 \pm 0.0393 $ & $ 81.92 \pm 0.0297 $ & $ 97.89 \pm 0.0066 $\\
		24       & $ 76.98 \pm 0.0398 $ & $ 87.91 \pm 0.0229 $ & $ 96.41 \pm 0.0091 $\\
		25       & $ 86.86 \pm 0.0330 $ & $ 83.90 \pm 0.0516 $ & $ 98.08 \pm 0.0016 $\\
		26       & $ 80.97 \pm 0.0549 $ & \colorbox{green}{$ 64.20 \pm 0.0280 $} & $ 97.83 \pm 0.0073 $\\
		27       & $ 87.70 \pm 0.0293 $ & $ 83.63 \pm 0.0528 $ & $ 95.09 \pm 0.0101 $\\
		28       & $ 86.65 \pm 0.0293 $ & $ 74.59 \pm 0.0522 $ & $ 98.27 \pm 0.0072 $\\
		29       & $ 78.24 \pm 0.0217 $ & $ 77.06 \pm 0.0418 $ & $ 96.02 \pm 0.0096 $\\
		30       & $ 85.41 \pm 0.0447 $ & $ 84.62 \pm 0.0402 $ & $ 98.15 \pm 0.0082 $\\
		31       & $ 91.34 \pm 0.0212 $ & $ 75.03 \pm 0.0745 $ & $ 96.12 \pm 0.0165$\\
		32       & $ 91.09 \pm 0.0222 $ & $ 89.85 \pm 0.0778 $ & $ 96.72 \pm 0.0088 $\\
		33       & $ 92.29 \pm 0.0237 $ & $ 80.02 \pm 0.0630 $ & $ 96.13 \pm 0.0112 $\\
		34       & $ 89.66 \pm 0.0414 $ & $ 77.84 \pm 0.0780 $ & $ 98.30 \pm 0.0104 $\\
		35       & $ 92.34 \pm 0.0545 $ & $ 78.19 \pm 0.0566 $ & $ 98.07 \pm 0.0070 $\\
		36       & $ 86.41 \pm 0.0395 $ & $ 81.87 \pm 0.0191 $ & $ 97.74 \pm 0.0064 $\\
		37       & $ 93.08 \pm 0.0162 $ & $ 85.93 \pm 0.0366 $ & $ 98.03 \pm 0.0031 $\\
		38       & $ 81.88 \pm 0.0343 $ & $ 84.56 \pm 0.0368 $ & $ 98.66 \pm 0.0035 $\\
		39       & $ 91.86 \pm 0.0208 $ & $ 83.74 \pm 0.0596 $ & $ 97.29 \pm 0.0048 $\\
		40       & $ 91.33 \pm 0.0212 $ & $ 91.30 \pm 0.0236 $ & $ 98.82 \pm 0.0040 $\\
		41       & $ 83.05 \pm 0.0694 $ & $ 79.63 \pm 0.0569 $ & $ 97.25 \pm 0.0090$  \\
		42       & \colorbox{green}{$ 70.94 \pm 0.0668 $} & $ 73.62 \pm 0.0285 $ & \colorbox{green}{$ 92.51 \pm 0.0152 $}\\
		43       & \colorbox{red}{$ 96.42 \pm 0.0093 $}& $ 85.22 \pm 0.0759 $ & $ 98.54 \pm 0.0046 $\\
		44       & $ 92.08 \pm 0.0264 $ & $ 87.49 \pm 0.0612 $ & $ 98.88 \pm 0.0043 $\\
		45       & $ 88.86 \pm 0.0204 $ & $ 79.56 \pm 0.0405 $ & $ 95.92 \pm 0.0111 $\\
		46       & $ 89.45 \pm 0.0170 $ & $ 71.50 \pm 0.0497 $ & $ 97.40 \pm 0.0073 $\\
		47       & $ 87.81 \pm 0.0238 $ & $ 77.32 \pm 0.0494 $ & $ 96.81 \pm 0.0063 $\\
		48       & $ 96.36 \pm 0.0118 $ & $ 88.44 \pm 0.0683 $ & $ 97.96 \pm 0.0072 $\\
		49       & $ 90.26 \pm 0.0490 $ & $ 79.96 \pm 0.0581 $ & $ 96.06 \pm 0.0248 $\\
		\hline
		Ave. & $ \boldsymbol{89.41} \pm 0.0584 $ & $ \boldsymbol{83.42} \pm 0.0677 $ & $ \boldsymbol{97.41} \pm 0.0132$\\
		\hline
		\hline
	\end{tabular}
\end{table}

\begin{figure}[t]
	\subfloat[SVM with a Gaussian kernel]{\includegraphics[width = \linewidth]{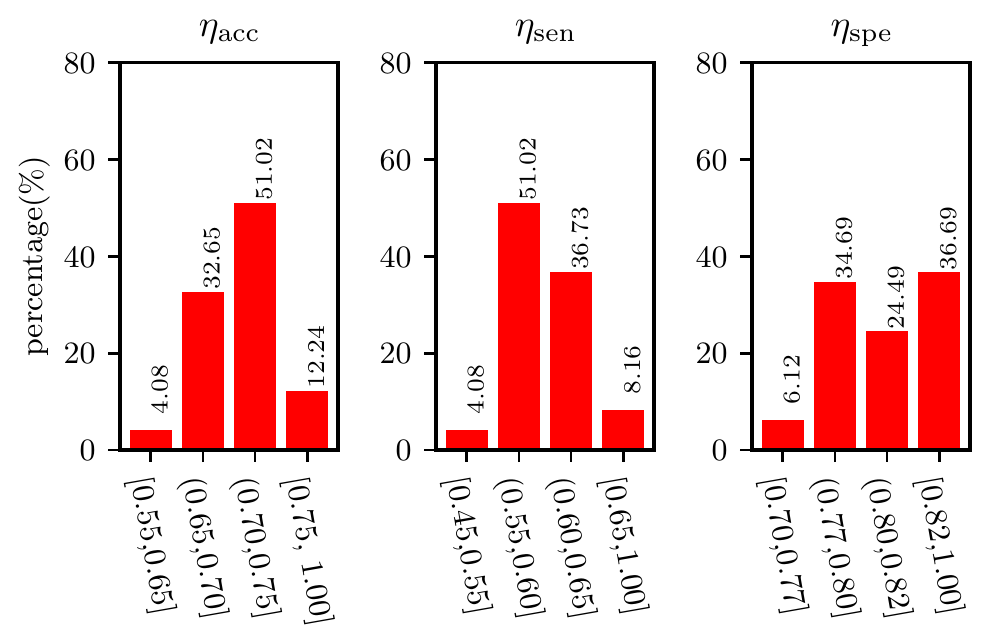}}\\
	\subfloat[SVM-BF with $ \gamma_{Br} = 0.9 $]{\includegraphics[width = \linewidth]{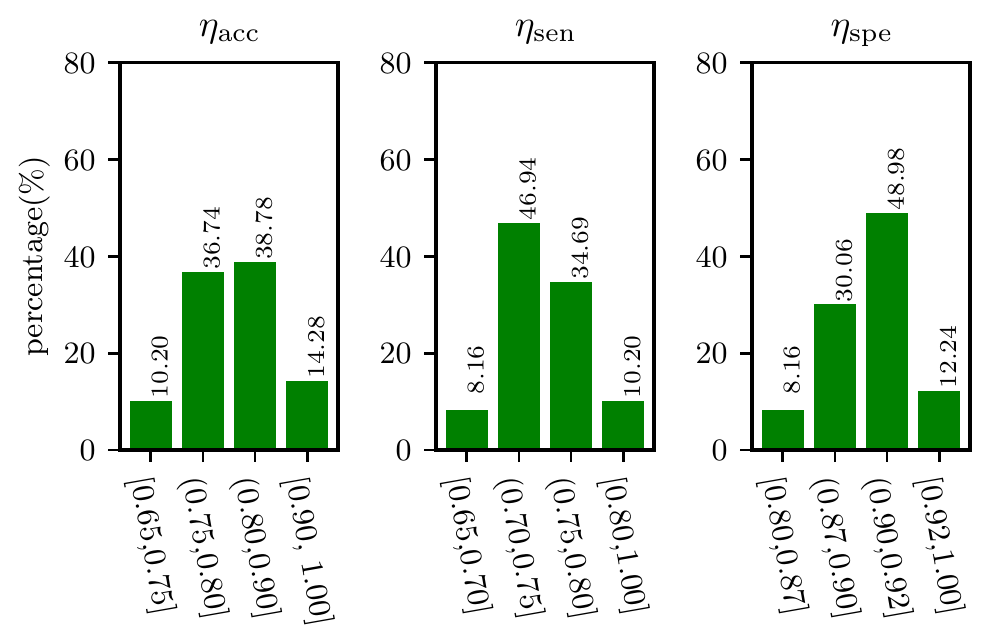}}\\
	\subfloat[GMM-HMM with $ Br^{\mathrm{c}} = 0.9 $]{\includegraphics[width = \linewidth]{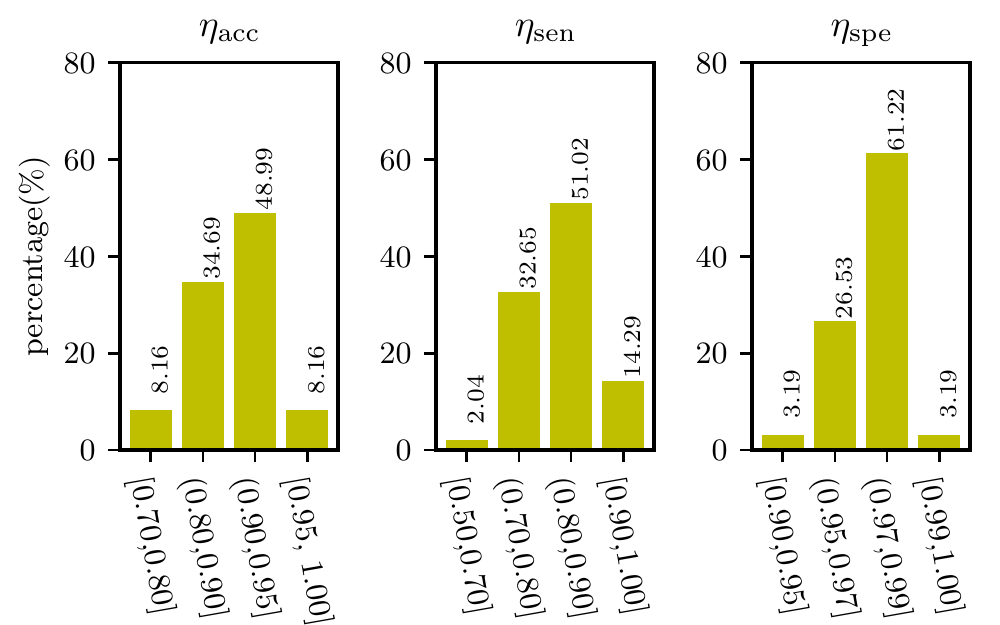}}\\
	\caption{Statistical results of performance metrics using (a) the SVM method, (b) the SVM-BF method, and (c) the GMM-HMM method.}
	\label{St_results}
\end{figure}

\begin{itemize}
	\item If the inferred output from (\ref{eq11}) is lager than a preset critical value $ Br^{\mathrm{c}} $, i.e., $ Br^{\mathrm{c}}  \leq \widehat{Br} $, we believe that the driver is braking. In addition, if $ \widehat{Br} > 1.0 $, it is believed that the driver is braking.
	\item If the inferred output from (\ref{eq11}) is smaller than $ Br^{\mathrm{c}} $, i.e., $ \widehat{Br} \leq Br^{\mathrm{c}} $, we believe that the driver is not braking.
\end{itemize}
Thus, we can decode the output results by

\begin{equation}\label{eq18}
\widehat{Br} = 
\begin{cases}
0, & \mathrm{if} \ \ \widehat{Br} \leq Br^{\mathrm{c}} \\
1, & \mathrm{otherwise}  \\
\end{cases}.
\end{equation}

Obviously, different critical values of $ Br^{\mathrm{c}} \in (0,1) $  could influence the model performance. We will discuss this in the following Section \textit{C. Result Analysis}. 

After decoding the experiment results using (\ref{eq18}), we can then compute the numbers of TP, TN, FP, and FN actions occurring in all of the test datasets. Fig. \ref{Example_res} shows an example of the inferred results of one fold test data from driver \#4 using the GMM-HMM method. We can see that, for driver \#4, the GMM-HMM can infer the braking actions with a high level of performance, specifically with an accuracy of 93.87\%, sensitivity of 83.76\%, and specificity of 98.27\%. In addition, Table \ref{Table_res} shows the means and standard deviations of ten cross-validation results regarding the three evaluation metrics for each driver participant. We know that the GMM-HMM method can achieve a remarkable performance, with accuracy of 96.42\%  (driver \#43), sensitivity of 95.41\% (driver \#9), and specificity of 99.54\% (driver \#8), as highlighted in red in Table \ref{Table_res}. Note that, for all driver participants, the GMM-HMM method is able to infer their braking actions, with an accuracy of 89.41\%, a sensitivity of 83.42\%, and a specificity of 97.41\% on average. However in some special cases, the performance is slightly lower than the average performance, as highlighted in green in Table \ref{Table_res}.

\begin{figure}[t]
	\centering
	\includegraphics[width = \linewidth]{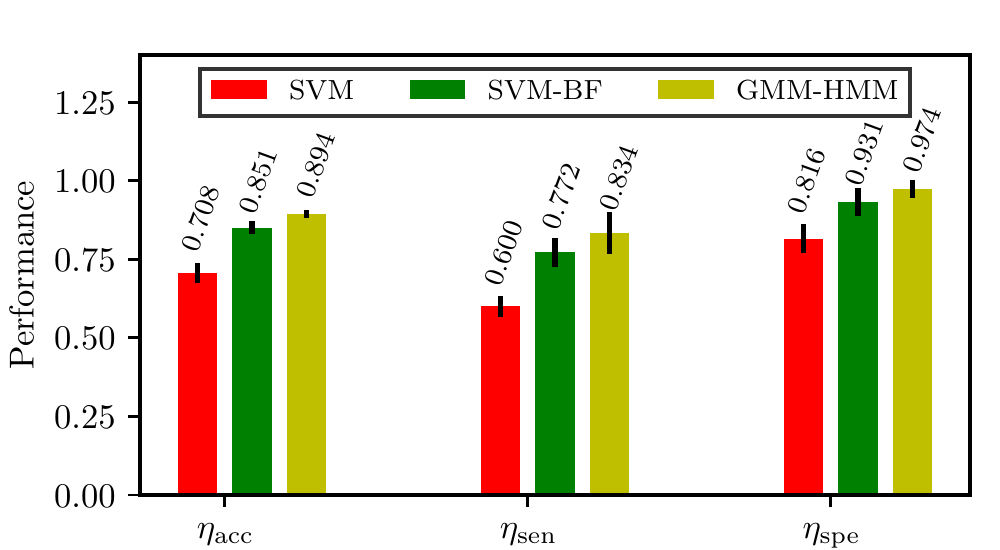}
	\caption{Comparison results of performance metrics using the SVM,  SVM-BF, and GMM-HMM methods with $ Br^{\mathrm{c}} = 0.9$ and $ \gamma_{Br} = 0.9 $.}
	\label{fig:comparison}
\end{figure}

\subsection{Results Analysis}
\subsubsection{Performance Analysis}
Table \ref{Table_res} shows the statistical performance metrics for each driver using the GMM-HMM. In order to show the benefits of GMM-HMM, we compare it with SVM and SVM-BF. Fig. \ref{St_results} shows the distributions of the mean values regarding the three performance metrics of all driver participants with different methods. In order to make comparison flexible, each performance metric (i.e., $ \eta_{\mathrm{acc}} $, $ \eta_{\mathrm{sen}} $, and $ \eta_{\mathrm{spe}} $) is further divided into four intervals for each method. The percentage of performance values in each interval is then computed as the ratio between the number of performance values located in this interval and the total number of driver participants. In each sub-figure of Fig. \ref{St_results}, the horizontal axis is the interval of performance metrics and the vertical axis is the percentage.

\begin{itemize}
	\item {\it Accuracy}: Comparing the left plots of Fig. \ref{St_results}(a)-(c), we can see that the GMM-HMM achieves a higher accuracy falling in the interval of $ [0.90, 0.95) $ with a percentage of 48.99\% than other two methods. More specifically, about half of the driver participants only achieve accuracy falling in $ [0.70, 0.75) $ for SVM and 38.78\% of the driver participants with accuracy falling in $ [0.80, 0.90) $ for SVM-BF. In addition, GMM-HMM obtains the accuracy of higher than 0.95 such as for driver \#1, driver \#14, driver \#43, and driver \#48, as shown in Table \ref{Table_res}. By using GMM-HMM, 34.69\% of drivers are able to obtain the accuracy falling in $ [0.80, 0.90) $ and only 8.16\% of drivers fall in a low accuracy range $ [0.70, 0.80) $.
	
	\item {\it Sensitivity}: From Table \ref{Table_res}, we can see that GMM-HMM achieves a slightly lower sensitivity of 83.42\% on average, but it performs a better sensitivity, compared to SVM and SVM-BF, as shown in the middle plots of Fig. \ref{St_results}(a)-(c). More specifically, when using GMM-HMM, more than 50\% and 30\% of the 49 drivers achieve the sensitivity falling in $ [0.80, 0.90) $ and $ [0.70, 0.80) $, respectively. In addition, only 2.04\% of the driver participants achieve the sensitivity of lower than 0.70, compared to more than 90\% of drivers with SVM and about 55 \% of drivers with SVM-BF.
	
	\item {\it Specificity}: From the right plots of Fig. \ref{St_results}(a)-(c), we found that the GMM-HMM obtains the highest level of specificity among the three methods. More specifically, 61.22\% of driver participants obtain a specificity falling in $ [0.97,0.99) $ and 26.53\% of all driver participants obtain a specificity of $ [0.95,0.97) $. In addition, Table \ref{Table_res} shows that the specificity is also more like to be greater than 0.99 for some drivers such as driver \#8, driver \#18 and driver \#22. The best value of specificity is 99.54\% (driver \#8) and the worst value of specificity is still over 92\% (driver \#42).  However, for the SVM and SVM-BF methods, both of them were not able to achieve such highlight performance.
\end{itemize}

In order to evaluate the average performance of different methods, Fig. \ref{fig:comparison} presents the statistical results of means and standard deviations regarding the three performance metrics over all drivers. We can see that the GMM-HMM achieves a better performance than the other two methods. Using GMM-HMM improves the performance by 26.37\% in accuracy, 39.06\% in sensitivity, 19.36\% in specificity with respect to those using SVM and by 5.05\% in accuracy, 8.03\% in sensitivity, 4.62\% with respect to those using SVM-BF.

\begin{figure}[t]
	\centering
	\includegraphics[width = \linewidth]{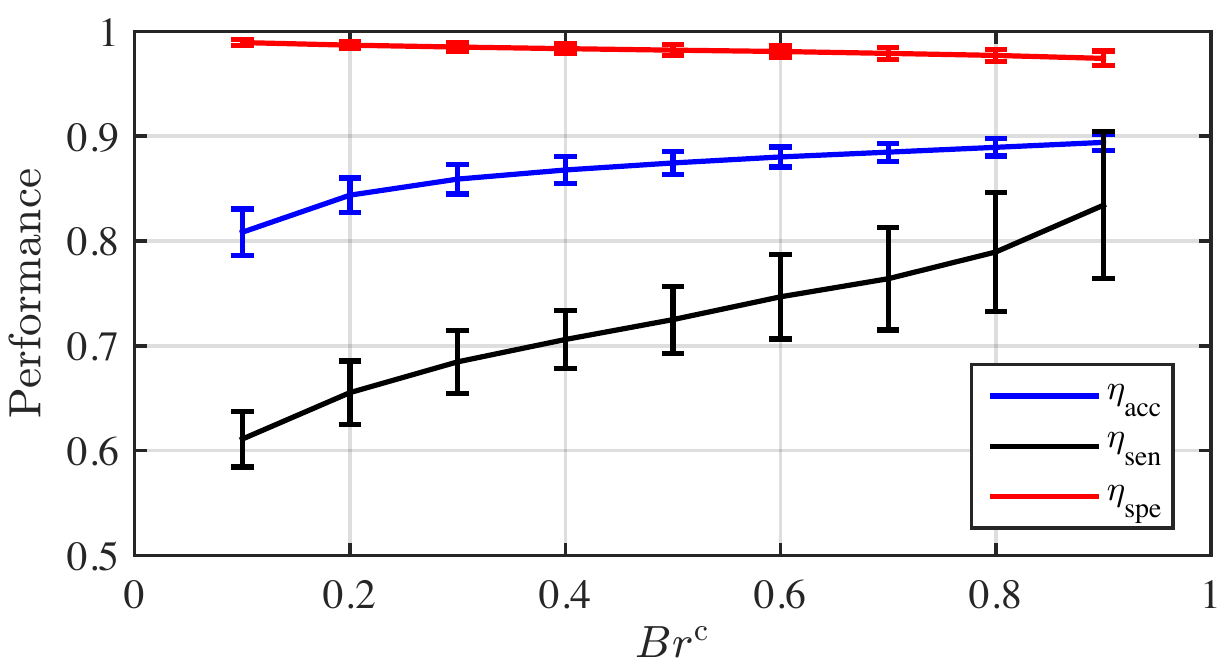}
	\caption{The influences of brake critical value, $ Br^{\mathrm{c}}$, on model performances.}
	\label{Br_cri}
\end{figure}


\subsubsection{Results Discussion}

From the above-mentioned analysis, we found that all of the three methods performed well in terms of accuracy and specificity on average, but less well in terms of sensitivity, as shown in Fig. \ref{fig:comparison}.

A high level of sensitivity indicates that GMM-HMM can correctly infer the case where the driver will brake when following a preceding vehicle, which can be used to design human-friendly FCW systems. For example, when the driver does brake and the inferred outcome is a braking action in the current driving situation, and then comparing the inferred results with the driver's real action can help the FCW systems to determine ``{\it Should I send a forward collision warning to the driver?}".

A high specificity value indicates that GMM-HMM can infer that the driver will not take braking actions. The results then are helpful to solve over-warning issues in the FCW systems. For example, if we correctly infer that the driver would not brake in the future situation as that driver usually does, we simply do not need to give a warning to the driver, thus avoiding the over-warning issues. 

\subsubsection{Influence of the Critical Value ($Br^{\mathrm{c}}$)}
Fig. \ref{Br_cri} shows the influences of the critical value on model performances when using GMM-HMM. We can conclude that, with an increasing critical value $ Br^{\mathrm{c}} $, 
\begin{itemize}
	\item the accuracy $ \eta_{\mathrm{acc}} $ improves slightly from 80.84\%  with $Br^{\mathrm{c}}$ = 0.1 to 89.41\%  with $Br^{\mathrm{c}}$ = 0.9;
	\item the sensitivity $ \eta_{\mathrm{sen}} $ improves significant about 22.30\%;
	\item the specificity $ \eta_{\mathrm{spe}} $ decreases slightly from 98.93\% with $Br^{\mathrm{c}}$ = 0.1 to 97.41\%  with  $Br^{\mathrm{c}}$ = 0.9.
\end{itemize}
In other words, the accuracy and specificity have smaller standard deviations of 0.0123 and 0.0049, respectively, compared to the standard deviation of sensitivity with 0.0402. Therefore, we selected $Br^{\mathrm{c}} = 0.9$ to make a trade-off between three performance metrics and show the results in Fig. \ref{Example_res} - Fig. \ref{fig:comparison} as well as in Table \ref{Table_res}.


\section{External Discussion and Future Work}
In this paper, the driver's braking action in car-following scenarios was learned and inferred using the developed GMM-HMM method and achieved a high level of accuracy, with 97.41\% on average. Here, we mainly focused on whether the driver will brake when following a preceding car, which was the first step in modeling driver braking behavior. Individuals' driving style of hitting the brake pedal were not included. Potential directions in future work are discussed as follows.

\subsection{Influence of Road Curvature}
Note that the influence of road profiles with a large road curvature were not included. In general, the road curvature will influence drivers' braking decision and action.  Researchers \cite{chandrasiri2016driving} demonstrated that drivers' braking action differs across individuals while approaching and negotiating a road segment with a large curvature, particularly at the beginning \cite{erseus2010driver}. For example, when entering curvy roads, some drivers tend to prefer braking on the straight road segment, far in advance of the curve beginning, to obtain the desired speed; conversely, others may tend to brake hardly at the beginning of the curve. Future work will take larger road curvatures into consideration when inferring the driver's braking action.

\subsection{Bounded Characteristics of Variables}
Most observed driving data from drivers usually have bounded support features\cite{wang2017evaluation,wang2017development}. For example, in the car-following scenario, the relative range between the ego and leading vehicle is usually larger than a critical value, and also drivers usually prefer certain relative ranges or vehicle speeds, thereby the distributions of the relative range and the vehicle speed will have the bounded supports. In the developed GMM-HMM method, the bounded features of driving data were not considered, which might be one of the factors that cause a large deviation in the metric of specificity. Therefore, we will develop a more robust and flexible Gaussian mixture model to fit all kinds of driving data in future work.

\subsection{Applications in Future Works}
This paper investigated on whether a driver will brake when following a vehicle. The inferred outcomes of drivers' braking action make the FCW systems acceptable for end-users, as discussed in Section V-\textit{C}. However, other characteristics of driver behavior such as the style of hitting brake pedal (e.g., hard and gentle) and kinematic characteristics \cite{wang2016development} of the ego and preceding vehicles should be included when designing human-friendly FCW systems. 

\section{Conclusions}
This paper has developed a GMM-HMM method for learning and inferring drivers' braking action in car-following scenarios. The driver's braking behavior was formulated from a perception-decision-action perspective and the driving situation was described using four variables: speed of the ego vehicle, relative distance and speed between the ego and preceding vehicle, and time to collision. The braking action was discretized into binary values (i.e., $ 1 $ - brake and $ 0 $ - no brake). The relationships between perceptions and brake actions were modeled using a joint distribution of multi-variable Gaussian regression functions.  The GMM-HMM method was validated using naturalistic driving data collected from 49 drivers. A series of comparative experiment were conducted among the SVM, SVM-BF and GMM-HMM methods. The experiment results shown that the GMM-HMM achieves the best performance, with an accuracy of 89.41\%, sensitivity of 83.42\%, and specificity of 97.41\% on average.

\appendices
\section{}
In this Appendix, the calculation of transfer matrix $ \mathcal{T} $ in Section III-{\it B} is presented. 
Given the training data set with $ n $ data points $ \boldsymbol{\zeta}_{t} $:
\begin{equation*}
\mathcal{S}_{Train} = \{ \boldsymbol{\zeta}_{1}, \boldsymbol{\zeta}_{2}, \cdots, \boldsymbol{\zeta}_{t}, \cdots, \boldsymbol{\zeta}_{n} \}.
\end{equation*}
For each data point $ \boldsymbol{\zeta}_{t}  $, we assume that $ \boldsymbol{\zeta}_{t}  $ is subject to the mode $ \mathcal{M}_{i} \in \{1, 2, \cdots, M\} $ if 

\begin{equation}
\mathcal{M}_{i} = \underset{i \in \{ 1,2, \cdots, M\} }{\max}  \ \mathcal{N}_{i}(\boldsymbol{\zeta}_{t}; \boldsymbol{\mu}_{i}, \Sigma_{i}).
\end{equation}
Therefore, each $ \boldsymbol{\zeta}_{t}  $ has a mode $ \mathcal{M}_{t} \in \{ 1, 2, \cdots, M \} $, and we obtain a mode sequence with the same number of training data points
\begin{equation}\label{appedix.1}
\{\mathcal{M}_{t}\}_{t=1}^{n} \Longleftrightarrow \mathcal{S}_{Train} = \{\boldsymbol{\zeta}_{t}\}_{t=1}^{n}. 
\end{equation}
The transfer probability between mode $ i $ and mode $ j $ can then be estimated by

\begin{equation}
T_{i,j} = \frac{F_{i,j}}{n_{i}}, \ i, j = 1, 2, \cdots, M
\end{equation}
where $ F_{i,j} $ is the frequency of transferring from mode $ \mathcal{M}_{i} $ to $ \mathcal{M}_{j} $, and $ n_{i} $ is the total number of training data points in mode $ \mathcal{M}_i $. In this research, $ n$ is larger than $5\times 10^{5} $. Finally, we obtain the transfer matrix
\begin{equation}
\mathcal{T} = 
\begin{bmatrix}
T_{1,1} & T_{1,2}  & \cdots & T_{1,M-1} & T_{1,M} \\ 
T_{2,1} & T_{2,2}  & \cdots & T_{2,M-1} & T_{2,M} \\ 
\vdots & \vdots   &  \ddots & \vdots  &  \vdots \\ 
T_{M,1} & T_{M,2}  & \cdots & T_{M,M-1} & T_{M,M}\\ 
\end{bmatrix}_{M\times M}.
\end{equation}

%
%

\ifCLASSOPTIONcaptionsoff
  \newpage
\fi



\bibliographystyle{IEEEtran}
\bibliography{Ref_bib}
%

%

\begin{IEEEbiography}[{\includegraphics[width=1in,height=1.25in,clip,keepaspectratio]{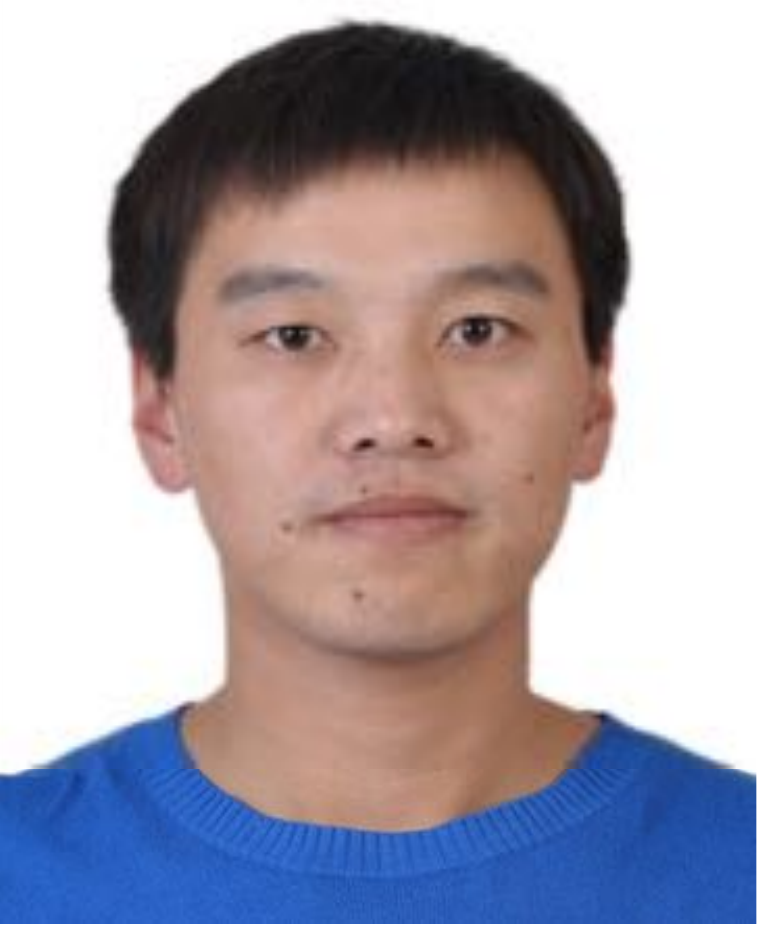}}]{Wenshuo Wang}
(S'15) received his B.S. in Transportation Engineering from ShanDong University of Technology,
Shandong, China, in 2012. He is a Ph.D. candidate for Mechanical Engineering, Beijing Institute of Technology (BIT). Now he is a visiting scholar studying in the School of Mechanical Engineering, University of California at Berkeley (UCB). Currently, he makes research under the supervisor of Prof. Junqiang Xi (BIT) and Prof. Karl Hedrick at Vehicle Dynamics \& Control Lab, University of California at Berkeley. His research interests include vehicle dynamics control, adaptive control, driver model, human-vehicle interaction, recognition and application of human driving characteristics. His work focuses on modeling and recognizing drivers behavior, making intelligent control systems between human driver and vehicle.
\end{IEEEbiography}

\begin{IEEEbiography}[{\includegraphics[width=1in,height=1.25in,clip,keepaspectratio]{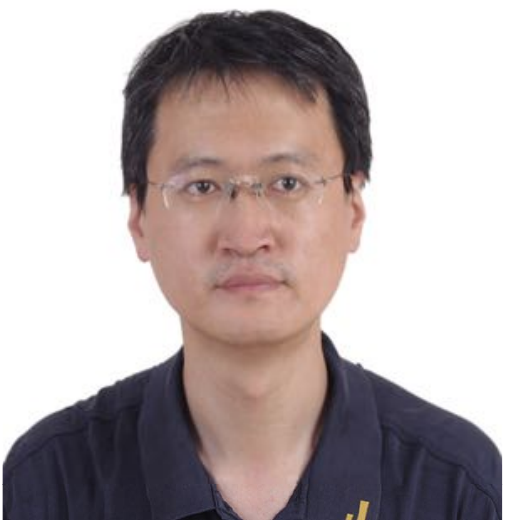}}]{Junqiang Xi} received the B.S. in Automotive Engineering from Harbin Institute of Technology, Harbin, China, in 1995 and the PhD in Vehicle Engineering from Beijing Institute of Technology (BIT), Beijing, China, in 2001. In 2001, he joined the State Key Laboratory of Vehicle Transmission, BIT. During 2012-2013, he made research as an advanced research scholar in Vehicle Dynamic and Control Laboratory, Ohio State University(OSU), USA. He is Professor and Director of Automotive Research Center in BIT currently. His research interests include vehicle dynamic and control, power-train control, mechanics, intelligent transportation system and intelligent vehicles.
\end{IEEEbiography}

\begin{IEEEbiography}[{\includegraphics[width=1in,height=1.25in,clip,keepaspectratio]{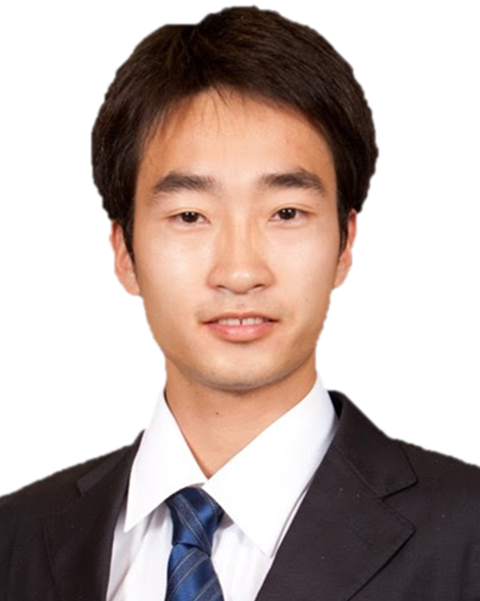}}]{Ding Zhao} received his Ph.D. degree in 2016 from the University of Michigan, Ann Arbor. He is currently an Assistant Research Scientist at Department of Mechanical Engineering, University of Michigan, Ann Arbor. His research interests include autonomous driving, connected/smart city, energy efficiency, human-machine interaction, cybersecurity, and big data analytics.
\end{IEEEbiography}

\end{document}